\documentclass{article}

\PassOptionsToPackage{numbers}{natbib}

\usepackage[preprint]{neurips_2024}
\makeatletter
\renewcommand{\@noticestring}{}
\makeatother

\usepackage[utf8]{inputenc} % allow utf-8 input
\usepackage[T1]{fontenc}    % use 8-bit T1 fonts
\usepackage{hyperref}
\hypersetup{colorlinks,allcolors=black}
\usepackage{url}            % simple URL typesetting
\usepackage{booktabs}       % professional-quality tables
\usepackage{amsfonts}       % blackboard math symbols
\usepackage{graphicx}       % for \resizebox
\usepackage{nicefrac}       % compact symbols for 1/2, etc.
\usepackage{microtype}      % microtypography
\usepackage{xcolor}         % colors
\usepackage{multirow}
\usepackage{array}
\usepackage{threeparttable}
\usepackage{tabularx}
\usepackage{amsmath}
\usepackage{tikz}
\usepackage{float}
\usepackage{placeins}
\usetikzlibrary{arrows.meta, positioning, shapes.geometric}
\title{AIM: Adversarial Information Masking for Faithfulness Evaluation of Saliency Maps}
\author{
Chia-Ying Hsieh \quad Hsin-Yuan Fang \quad Chun-Shu Wei\\
National Yang Ming Chiao Tung University
}

\begin{document}

\maketitle

\begin{abstract}
Post-hoc saliency methods are widely used to interpret deep neural networks, but their faithfulness is difficult to evaluate reliably. Existing evaluations mask features according to saliency-induced feature ordering and measure performance degradation, but this degradation can be confounded by the masking operator: zero masking may create out-of-distribution artifacts, while interpolation-based masking may preserve residual predictive information. We propose Adversarial Information Masking (AIM), a saliency-guided adversarial feature replacement framework for evaluating both saliency-map faithfulness and masking-operator reliability. AIM replaces selected features with values from an adversarial counterpart of the input and compares degradation under complementary masking orders. We assess reliability using random-attribution bias and stability of explanation-method faithfulness rankings. Experiments on image, audio, and EEG tasks suggest that AIM reduces masking-induced bias compared with zero and interpolation-based masking, while revealing modality-dependent differences between signed and unsigned attributions.
\end{abstract}

\section{Introduction}

Post-hoc explanation methods are widely used to interpret deep neural networks, whose internal decision processes are often difficult to inspect directly  \cite{zeiler2014visualizing, samek2016evaluating, lundberg2017unified, ancona2017towards}. These methods support model debugging, evaluation of AI-assisted decision making, and discovery of informative patterns that may not be obvious to human observers  \cite{goodman2017european, cadamuro2016assessing, adebayo2020sanity, shrikumar2017learning}. Among different explanation paradigms, feature attribution methods assign importance scores to input features and are commonly used to produce saliency maps  \cite{adebayo2022post}. They have been applied in computer vision, audio analysis, and neural signal decoding, including electroencephalography (EEG), where deep learning models are increasingly used to decode structured brain signals  \cite{roy2019deep, tjoa2020survey, pan2022explaining, bilodeau2024review}.

A central question is whether a saliency map is \emph{faithful}: do the highlighted features genuinely reflect the information used by the model? A common evaluation strategy masks features according to the feature ordering induced by a saliency map and measures the resulting change in model performance. Under this principle, features with larger attribution scores are expected to be more relevant to the model prediction  \cite{shah2021hiding}; masking highly ranked features should therefore degrade performance more than masking low-ranked features. This idea connects fidelity-oriented evaluation  \cite{yeh2019fidelity} and robustness-oriented evaluation  \cite{hsieh2020evaluations, fang2024evaluating}, both of which test whether attribution scores are aligned with model behavior.

However, the measured degradation is not determined only by the explanation method. It can also be strongly affected by the masking operator. Zero masking may create out-of-distribution artifacts, especially when zeros are not natural values for the input domain. Interpolation-based masking can reduce some artifacts, but it relies on hand-crafted imputation rules and may leak information through the shape, location, or reconstructed content of the masked region. As a result, a faithfulness score may reflect masking-induced bias rather than true model reliance on the features identified by the saliency map.

This issue is particularly important across modalities. In images, masking may create artificial patches; in audio, it may introduce silence or time--frequency discontinuities; and in EEG, it may disrupt temporal structure, spectral content, or spatial covariance. A reliable evaluation framework should therefore test not only the explanation method, but also the masking method used to evaluate it.

We propose \textbf{Adversarial Information Masking (AIM)}, which uses a PGD-generated adversarial counterpart as a saliency-guided feature-replacement source. Rather than treating adversarial perturbation as a global robustness probe, AIM selectively copies adversarial values into saliency-defined feature subsets within the same MoRF/LeRF protocol used for zeroing and interpolation. This design supports direct comparison among masking operators and enables two reliability diagnostics: random-attribution bias and stability of explanation-method rankings under complementary masking orders.

Our contributions are summarized as follows:
\begin{itemize}
    \item We formulate masking-induced bias as a key confounder in saliency faithfulness evaluation, where performance degradation may reflect artifacts of the masking operator rather than the quality of the explanation.
    \item We propose AIM, a saliency-guided adversarial feature replacement framework that constructs a full adversarial counterpart and selectively imputes adversarial perturbed inputs into saliency-defined feature subsets.
    \item We compare zero masking, interpolation-based masking, and AIM under a shared reliability evaluation protocol across image, audio, and EEG classification tasks based on random-attribution bias and the consistency of explanation-method rankings under complementary MoRF and LeRF masking orders.
    \item We analyze how signed and unsigned attributions affect faithfulness evaluation in structured signal data, highlighting that attribution sign and magnitude are not interchangeable across temporal and spectral domains.
\end{itemize}

\begin{figure}[t!]
\centering
\includegraphics[width=1\linewidth]{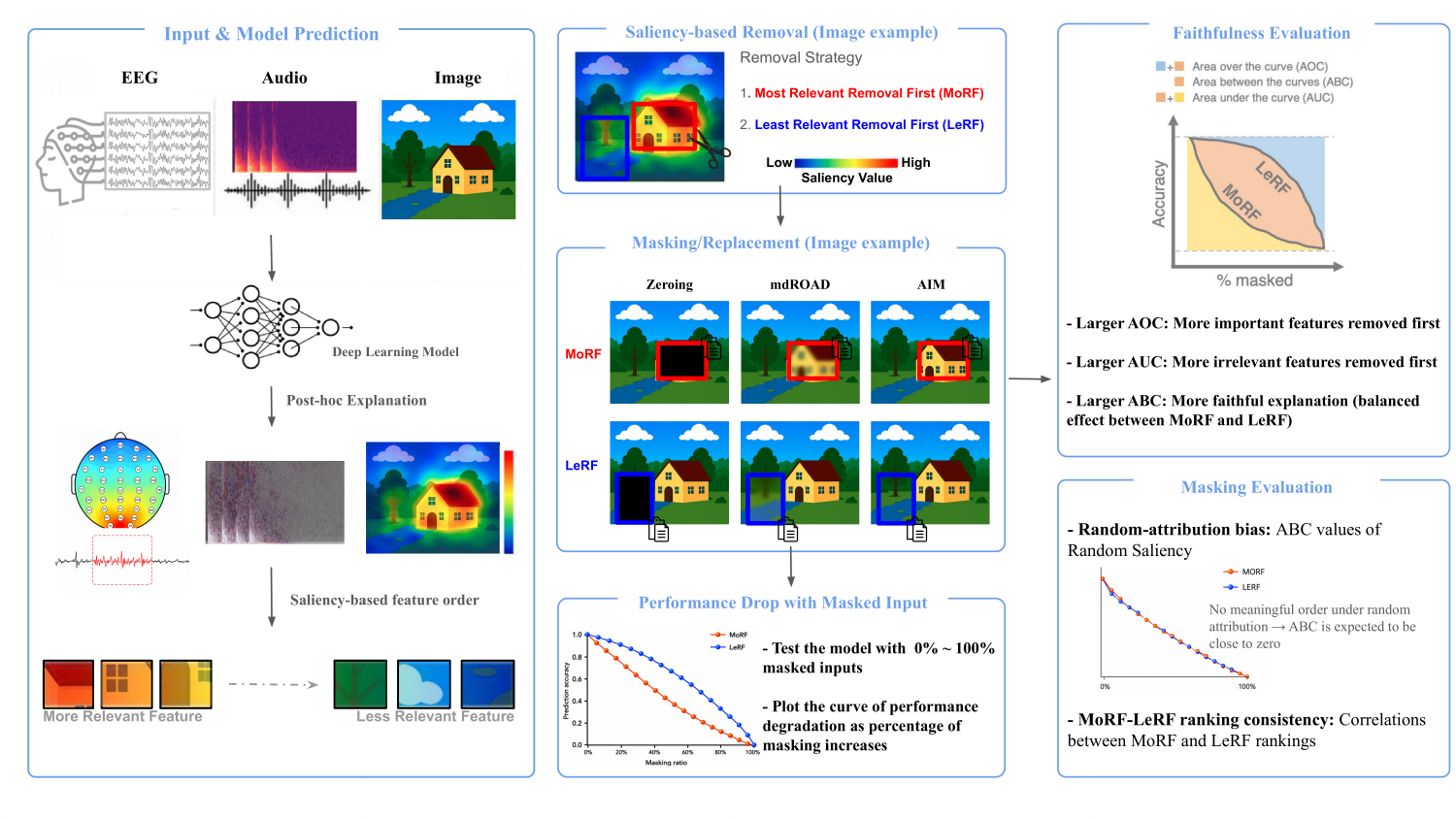}
\caption{
Overview of the proposed AIM framework. A post-hoc explanation method first produces a saliency map that induces a feature ordering. AIM then replaces selected features with their adversarial counterparts under complementary masking orders and evaluates faithfulness by comparing the resulting prediction degradation curves. The framework also assesses masking reliability through random-attribution bias and the consistency of explanation method faithfulness rankings.
}
\label{aim_framework}
\end{figure}

\section{Related Work}

We review explanation-quality evaluation with emphasis on masking-based faithfulness and masking-induced bias. A broader terminology summary is provided in Appendix~\ref{app:literature_table}.

\paragraph{Explanation Quality Criteria.}
Explanation quality has been studied through axioms, robustness, and faithfulness. Axiomatic work defines properties such as completeness, sensitivity, linearity, local accuracy, missingness, and consistency \cite{ancona2017towards, sundararajan2017axiomatic, shrikumar2017learning, lundberg2017unified, kindermans2019reliability}. Robustness studies examine stability under model randomization, input perturbation, or similar generation settings \cite{adebayo2018sanity, yeh2019fidelity, ravindran2023empirical}. Faithfulness asks whether explanations reflect the model's actual decision process, which is the focus of this work.

\paragraph{Masking-Induced Bias.}
Faithfulness is often evaluated by masking features according to saliency-induced feature ordering. However, masking can confound the result: fixed-value removal may create out-of-distribution inputs \cite{dabkowski2017real, hooker2018evaluating}, while mask patterns may leak class-relevant information \cite{rong2022consistent}. These effects can cause MoRF-LeRF ranking inconsistency \cite{tomsett2020sanity, rong2022consistent} and reduce reliability across datasets, models, and metrics \cite{tomsett2020sanity, rong2022consistent, brocki2022fidelity}. Thus, the masking method itself should be evaluated.

\paragraph{Removal and Imputation Frameworks.}
ROAR reduces removal bias by retraining models after fixed-value feature removal \cite{hooker2018evaluating}. DiffROAR explores diffusion-based replacement \cite{shah2021hiding}. ROAD addresses leakage and ranking inconsistency through noisy interpolation \cite{rong2022consistent}, while GOAR introduces geometric and diffusion-based purification \cite{park2023geometric}. Corruption-and-training designs have also been used for time-series explanations \cite{turbe2023evaluation}. These methods are valuable but often require retraining, imputation rules, or auxiliary models.

\paragraph{Adversarial Perturbation for Evaluation.}
Adversarial perturbation provides another evaluation route, as deep networks are sensitive to small adversarial changes \cite{goodfellow2014explaining}, including EEG models \cite{zhang2019vulnerability}. Robustness-$S$ evaluates whether important features are more adversarially sensitive \cite{hsieh2020evaluations}, and OAR adds OOD-aware reweighting \cite{fang2024evaluating}. Nieradzik et al. showed that insertion/deletion metrics for CNN attribution maps suffer from distribution shift and proposed adversarial perturbation instead of pixel modification \cite{nieradzik2025reliable}. AIM builds on this idea, but uses adversarial counterparts as saliency-guided feature replacement sources within a MoRF/LeRF protocol, enabling comparison with zeroing and interpolation-based masking.

\paragraph{Cross-Modal Evaluation.}
Cross-modal faithfulness evaluation remains limited. Prior EEG work evaluates spatial, temporal, and spectral attribution \cite{apicella2022toward, cui2023towards}, while ROAR-style and synthetic-signal studies have been used in neural decoding \cite{torres2023evaluation, ravindran2023empirical}. Audio attribution has been studied with waveform or spectrogram masking \cite{becker2024audiomnist}. Recent studies highlight metric dependence \cite{gomez2022metrics, skliarov2025comparative}, and F-Fidelity addresses OOD and leakage through explanation-agnostic fine-tuning and random masking across modalities \cite{zheng2025ffidelity}. Existing methods still often rely on modality-specific perturbations, retraining, auxiliary models, or synthetic assumptions \cite{singh2021towards, rajpura2024explainable}. AIM addresses this gap through modality-general adversarial feature replacement and explicit masking-bias diagnostics.

\section{Method}

We evaluate post-hoc explanation methods by progressively masking input features according to the feature ordering induced by each saliency map. The framework compares complementary masking orders: Most Relevant First (MoRF), which masks high-importance features first, and Least Relevant First (LeRF), which masks low-importance features first. A faithful explanation should produce a larger performance drop under MoRF than under LeRF.

Given an input $x$, label $y$, trained model $f$, and saliency map $S$, we construct feature subsets $\Phi_k^M$ and $\Phi_k^L$ at masking step $k$, corresponding to the top- and bottom-ranked features under $S$. Applying a masking operator $\mathcal{T}$ produces two perturbed inputs:
\[
x_k^M = \mathcal{T}(x, \Phi_k^M), \qquad
x_k^L = \mathcal{T}(x, \Phi_k^L).
\]
The model is then evaluated on the perturbed inputs to obtain MoRF and LeRF performance curves. The overall pipeline is illustrated in Figure~\ref{aim_framework}.

We apply the same evaluation principle to image, audio, and EEG inputs, with features defined as spatial regions, waveform or spectrogram elements, and spatial, temporal, or spectral EEG components.

\subsection{Baseline Masking Strategies}

We consider two baseline masking strategies: \textbf{Zeroing} and \textbf{mdROAD}. These methods differ in how masked features are replaced. The comparison excludes GOAR \cite{park2023geometric} because its diffusion-based purification requires modality-specific auxiliary generative models and would confound the comparison of masking operators.

\paragraph{Zeroing.} directly removes selected features by setting them to a constant value:
\[
x'_{i \in \Phi} = 0,
\]
where $\Phi$ denotes the set of masked feature indices.

\paragraph{mdROAD.} extends the ROAD framework  \cite{rong2022consistent} to multiple domains by performing in-distribution imputation. Instead of removing features, masked regions are reconstructed using neighboring information to preserve local structure:
\[
x'_{i \in \Phi} \leftarrow \mathcal{I}(x_{i \notin \Phi}),
\]
where $\mathcal{I}(\cdot)$ denotes a domain-specific interpolation function.

While mdROAD aims to maintain realistic input statistics, the in-distribution reconstruction may retain residual class-relevant information, which can affect the reliability of faithfulness evaluation. Detailed formulations for different domains are provided in Appendix.

\subsection{Adversarial Information Masking (AIM)}

AIM replaces selected features with values from a PGD-generated adversarial counterpart, avoiding fixed constants and hand-crafted interpolation rules.

We first generate an adversarial counterpart $x^{\mathrm{adv}}$ using Projected Gradient Descent (PGD)  \cite{madry2017towards}:
\[
x^{\mathrm{adv}}_0 = x + \delta_0, \qquad \delta_0 \sim \mathcal{U}(-\epsilon, \epsilon),
\]
\[
x^{\mathrm{adv}}_{t+1}
= \Pi_{\mathcal{B}_\epsilon(x)}\left(
 x^{\mathrm{adv}}_t
 + \alpha \cdot \mathrm{sign}\left(
 \nabla_x \mathcal{L}(f(x^{\mathrm{adv}}_t), y)
 \right)
\right),
\]
where $\mathcal{L}$ is the cross-entropy loss, $\alpha$ is the step size, and $\Pi_{\mathcal{B}_\epsilon(x)}$ projects the perturbed input back to an $\epsilon$-bounded neighborhood around $x$.

For a selected feature set $\Phi$, AIM constructs the masked input by replacing the selected features with their adversarial counterparts:
\[
x'_{i \in \Phi} \leftarrow x^{\mathrm{adv}}_{i \in \Phi},
\]
or equivalently,
\[
x' = (1-M) \odot x + M \odot x^{\mathrm{adv}},
\]
where $M$ is a binary mask indicating the selected features.

AIM uses a model-aware adversarial counterpart to perturb selected features in a loss-increasing direction. Comparing MoRF and LeRF curves then tests whether highly ranked features affect model performance more than low-ranked features.

\subsection{Explanation Faithfulness Measurements}

Faithfulness is evaluated by measuring how model accuracy changes as increasingly large feature subsets are masked. A faithful ranking should cause a sharper degradation under MoRF than under LeRF  \cite{hooker2019benchmark}. Following area-based evaluation practice  \cite{tomsett2020sanity, apicella2022towards, brocki2022evaluation, cui2023explaining}, we compute three normalized metrics: Area Over Curve (AOC), Area Between Curves (ABC), and Area Under Curve (AUC):

\[
\mathrm{AOC} = \frac{1}{K} \sum_{k=1}^{K} 
\frac{\mathrm{Acc}(x_0) - \mathrm{Acc}(x_k^M)}
{\mathrm{Acc}(x_0) - \mathrm{Acc}(x_{\text{all masked}})},
\]

\[
\mathrm{ABC} = \frac{1}{K} \sum_{k=1}^{K} 
\frac{\mathrm{Acc}(x_k^L) - \mathrm{Acc}(x_k^M)}
{\mathrm{Acc}(x_0) - \mathrm{Acc}(x_{\text{all masked}})},
\]

\[
\mathrm{AUC} = \frac{1}{K} \sum_{k=1}^{K} 
\frac{\mathrm{Acc}(x_k^L) - \mathrm{Acc}(x_{\text{all masked}})}
{\mathrm{Acc}(x_0) - \mathrm{Acc}(x_{\text{all masked}})},
\]

where $x_k^M$ ($x_k^L$) denotes the input with the top-$k$ most (least) important features masked, respectively. Here, $k = 1, \ldots, K$, corresponding to increasing masking ratios (e.g., $5\%, 10\%, \ldots, 50\%$).  $\mathrm{Acc}(x)$ denotes the accuracy of the model on input $x$, and $x_{\text{all masked}}$ represents the fully masked input (i.e., all features are replaced by the masking strategy).

\subsection{Masking Reliability Evaluation}

We also evaluate masking-operator reliability using random-attribution bias, cross-configuration stability, and MoRF-LeRF ranking consistency.

\paragraph{Random-attribution bias:}
Random attribution does not encode meaningful feature importance. Therefore, the MoRF and LeRF curves induced by random attribution are expected to overlap across masking ratios. We use the ABC value under random attribution as a sanity check:
\[
\mathbb{E}\left[\mathrm{Acc}(x_k^L) - \mathrm{Acc}(x_k^M)\right] = 0.
\]
A random ABC value close to zero indicates that the masking operator does not assign artificial faithfulness to an uninformative explanation. A large random ABC suggests masking-induced bias, ordering asymmetry, or residual predictive information introduced by the masking process.

\paragraph{Stability across datasets and models:}
We analyze the standard deviation of AOC, ABC, and AUC across datasets and models. Lower variability indicates that the masking operator produces more stable faithfulness assessments across different data distributions and architectures.

\paragraph{MoRF-LeRF ranking consistency:}
We also examine whether a masking method produces stable explanation method faithfulness rankings under complementary feature-masking orders. For each masking ratio, explanation methods are ranked according to their performance degradation under MoRF and LeRF. We compute Spearman's rank correlation between the two rankings and average the correlation across masking ratios up to 50\%:
\[
\rho_{R_M, R_L} = 
\frac{\mathrm{cov}(R_M, R_L)}
{\sigma_{R_M} \, \sigma_{R_L}},
\]
where $R_M$ and $R_L$ denote the method-level rankings under MoRF and LeRF, respectively. Higher correlation indicates more stable method rankings under the chosen masking operator, although stability alone does not imply unbiased faithfulness evaluation.

\subsection{Experimental Setup}

We evaluate the proposed framework across selected image, audio, and EEG classification tasks. These tasks cover different feature structures, including spatial regions, temporal signals, time--frequency representations, and neural spatial--temporal--spectral patterns. The datasets are summarized in Table~\ref{tab:exp_setup}. Additional details on model architectures, modality-specific settings, preprocessing, and training protocols are provided in Appendix~\ref{app:experimental_setup_details}.

\begin{table}[t]
\centering
\scriptsize
\setlength{\tabcolsep}{3pt}
\renewcommand{\arraystretch}{0.3}
\caption{Datasets used in the cross-modal evaluation. The selected tasks cover image, audio, and EEG data with different feature structures, enabling evaluation of masking reliability across spatial, temporal, spectral, and time--frequency representations.}
\label{tab:exp_setup}

\begin{tabular}{c|c|c}
\toprule
\textbf{Modality} & \textbf{Dataset} & \textbf{Description} \\
\midrule

EEG 
& SMR (BCI IV 2A \cite{brunner2008bci}) 
& Motor imagery (time-asynchronous) \\

& ERN (BCI Challenge \cite{mattout2014bci}) 
& Event-related responses (time-locked) \\

& SSVEP (MAMEM \cite{martinez2007mamem}) 
& Frequency-locked signals (periodic stimuli) \\

\midrule

Audio 
& AudioMNIST \cite{becker2024audiomnist} 
& Spoken digits (structured speech) \\

& ESC-50 \cite{piczak2015dataset} 
& Environmental sounds (high variability) \\

& MSoS \cite{kroos2019msos} 
& Complex acoustic scenes \\

\midrule

Image 
& OxfordPet \cite{parkhi2012cats} 
& Fine-grained object recognition \\

& ImageNet \cite{deng2009imagenet} 
& Large-scale natural images \\

& BrainMRI \cite{bhuvaji2020brain} 
& Medical imaging (texture-based) \\

\bottomrule
\end{tabular}
\end{table}

\subsection{Post-hoc Methods for Saliency-based Feature Attribution}

We evaluate commonly used post-hoc explanation methods under a unified protocol. For EEG and audio, we use Gradient (GD)  \cite{simonyan2013deep}, Gradient$\times$Input (GI)  \cite{shrikumar2017learning}, SmoothGrad (SG)  \cite{smilkov2017smoothgrad}, SmoothGrad-Squared (SS)  \cite{hooker2019benchmark}, VarGrad (VG)  \cite{adebayo2018sanity}, and Integrated Gradients (IG)  \cite{sundararajan2017axiomatic}. For methods that naturally produce signed attribution values, we also evaluate their absolute-value variants: GDA, GIA, SGA, and IGA. This distinction follows prior discussions on whether attribution sign carries class-relevant information  \cite{bach2015pixel, ancona2018towards}. Random attribution is included as a sanity-check baseline.

For image data, we evaluate gradient-based methods together with common class activation mapping (CAM)-based methods: Grad-CAM (GC), Grad-CAM++ (GC++)  \cite{selvaraju2017grad, chattopadhyay2018grad}, Score-CAM (SC)  \cite{9150840}, and SmoothGrad-CAM++ (SGC++)  \cite{omeiza2019smooth}. Image saliency maps are aggregated across RGB channels using absolute values, so signed and unsigned variants are not separately analyzed for image explanations. Implementation details are provided in the appendix.

\section{Results}

We present the results in four parts. First, we compare faithfulness scores under AIM, interpolation-based masking, and zero masking across image, audio, and EEG tasks. Second, we evaluate masking reliability using random-attribution bias, MoRF-LeRF ranking consistency, and metric stability. Third, we analyze signed and unsigned attribution behavior in structured signal domains. Finally, we provide qualitative examples to examine whether the selected explanations align with model-relevant structures.

\subsection{Faithfulness Scores Across Modalities}

The tables \ref{tab:full_image}-\ref{tab:full_eeg_spectral} in the Appendix report the full AOC, ABC, and AUC results across modalities, feature domains, masking methods, models, and explanation methods. Table~\ref{tab:main_summary} summarizes the explanation method selected as most faithful by each masking framework.

The selected methods differ across masking operators, indicating that faithfulness conclusions are sensitive to how selected features are replaced. This supports our central motivation that masking should not be treated as a neutral step in saliency evaluation. Across many settings, the Random baseline yields low ABC values, as expected for uninformative feature orderings. However, the magnitude of random-attribution ABC varies across masking operators, motivating the masking-reliability analyses below.

\begin{table}[b!]
\centering
\scriptsize
\renewcommand{\arraystretch}{0.65}
\begin{tabular}{lcccc}
\toprule
\textbf{Modality / Domain} & \textbf{AIM} & \textbf{mdROAD} & \textbf{Zeroing} & \textbf{Full results} \\
\midrule
Image & SGC++ & GC & GC & Table~\ref{tab:full_image} \\
Audio waveform & GDA & IGA & IGA & Table~\ref{tab:full_audio_wave} \\
Audio spectrogram & SGA & GD & SG & Table~\ref{tab:full_audio_spec} \\
EEG spatial & SS & IG & IG & Table~\ref{tab:full_eeg_spatial} \\
EEG temporal & SS & GIA & IG & Table~\ref{tab:full_eeg_temporal} \\
EEG spectral & GD & SG & SG & Table~\ref{tab:full_eeg_spectral} \\
\bottomrule
\end{tabular}
\caption{
Summary of the explanation method selected as most faithful by each masking framework for each modality and feature domain. Full AOC, ABC, and AUC results are provided in Appendix~\ref{app:full_results}.
}
\label{tab:main_summary}
\end{table}

\subsection{Masking Reliability}

\paragraph{Random-attribution bias.}
Random attribution should not provide a meaningful feature ordering. Therefore, after averaging over random permutations, its MoRF and LeRF curves should remain close, and the corresponding ABC value should approach zero. A large random ABC suggests that the masking method itself may create artificial MoRF-LeRF separation, independent of explanation quality.

As shown in the Tables \ref{tab:full_image}-\ref{tab:full_eeg_spectral} across the evaluated settings, AIM generally yields random-attribution ABC values closer to zero than interpolation-based masking and zero masking. This suggests that AIM introduces less masking-induced bias under uninformative feature orderings.

\paragraph{MoRF-LeRF ranking consistency.}
Table~\ref{tab:consistency_summary} summarizes the number of configurations in which each masking method achieves the highest Spearman correlation between explanation method faithfulness rankings obtained from complementary feature-masking orders. The complete consistency results are provided in Appendix~\ref{app:consistency_results}.
AIM achieves the highest consistency in 30 out of 48 configurations, especially in audio and EEG settings. We interpret this result together with random-attribution bias and full-masking behavior, because high consistency alone does not guarantee unbiased masking. For example, interpolation-based masking can yield high consistency when residual predictive structure remains after masking.

\begin{table}[t]
\centering
\scriptsize
\setlength{\tabcolsep}{4pt}
\renewcommand{\arraystretch}{0.8}
\begin{tabular}{lcccc}
\toprule
\textbf{Modality} & \textbf{\# Configs.} & \textbf{AIM} & \textbf{mdROAD} & \textbf{Zeroing} \\
\midrule
Image & 9  & 5  & 4 & 0 \\
Audio & 12 & 8  & 2 & 2 \\
EEG   & 27 & 17 & 4 & 6 \\
\midrule
Overall & 48 & 30 & 10 & 8 \\
\bottomrule
\end{tabular}
\caption{
MoRF-LeRF ranking consistency summary. Entries indicate the number of dataset--model--domain configurations in which each masking method achieves the highest Spearman correlation between explanation method faithfulness rankings. Full results are provided in Appendix~\ref{app:consistency_results}.
}
\label{tab:consistency_summary}
\end{table}

\paragraph{Metric stability.}
AOC, ABC, and AUC capture related but non-identical aspects of MoRF and LeRF curves. AOC emphasizes degradation under high-importance masking, AUC emphasizes preservation under low-importance masking, and ABC summarizes their separation. We therefore use cross-metric agreement as a supporting diagnostic rather than as proof of faithfulness.
In our experiments, AIM tends to produce more consistent method-level conclusions across AOC, ABC, and AUC than zero masking and interpolation-based masking. Under baseline masking strategies, different metrics sometimes favor different explanation methods, suggesting that artifacts, residual information, or curve-shape differences can influence the final faithfulness ranking.

\subsection{Qualitative Framework Evaluation}

\begin{figure}[hbtp]
\centering
\includegraphics[width=\linewidth]{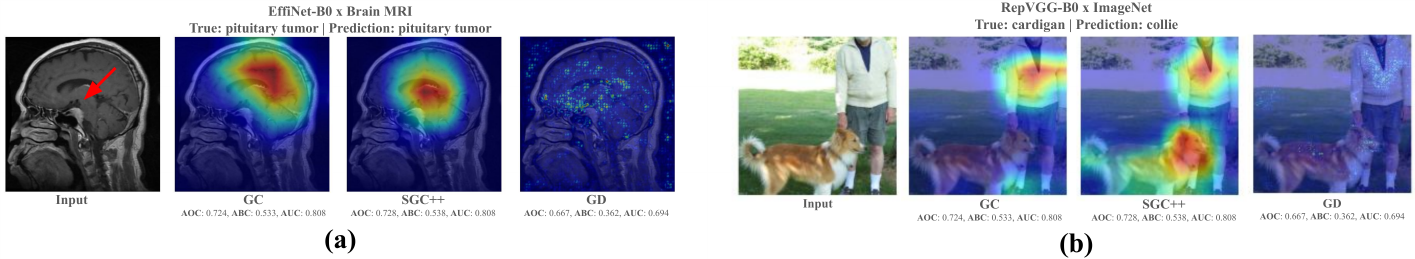}
\caption{
Image-domain examples. AIM selects SGC++, while interpolation-based masking and zero masking select GC; GD is shown as AIM's least faithful method. (a) Correct Brain Tumor MRI prediction using EfficientNet-B0; the red arrow indicates the pituitary region. (b) Incorrect ImageNet prediction using RepVGG-B0, where the model predicts \textit{collie} instead of the true class \textit{cardigan}.
}
\label{fig:qual_image}
\end{figure}

\paragraph{Image.}
Figure~\ref{fig:qual_image} compares explanations selected by different masking methods. AIM selects SGC++, whereas interpolation-based masking and zero masking select GC; GD is shown as AIM's least faithful method. In the Brain MRI example, SGC++ highlights a region closer to the pituitary area, while GC is more dispersed and GD emphasizes background regions. In the ImageNet error case, SGC++ better follows the dog region associated with the model's incorrect \textit{collie} prediction. These examples suggest that AIM-selected explanations can better reflect model-relevant structures in these cases.

\begin{figure}[hbtp]
\centering

\begin{minipage}[t]{0.44\linewidth}
\centering
\includegraphics[
    width=0.8\linewidth,
    keepaspectratio
]{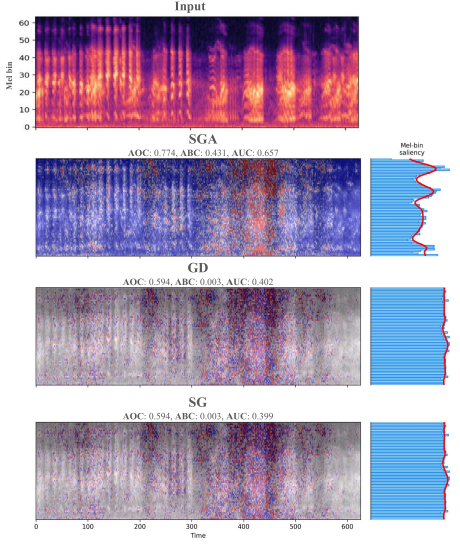}
\caption{
Audio spectrogram examples. AIM selects SGA, while interpolation-based masking selects GD and zero masking selects SG. Aggregated mel-bin profiles show frequency-selective saliency patterns.
}
\label{fig:qual_audio}
\end{minipage}
\hfill
\begin{minipage}[t]{0.54\linewidth}
\centering
\includegraphics[
    width=\linewidth,
    height=0.28\textheight,
    keepaspectratio
]{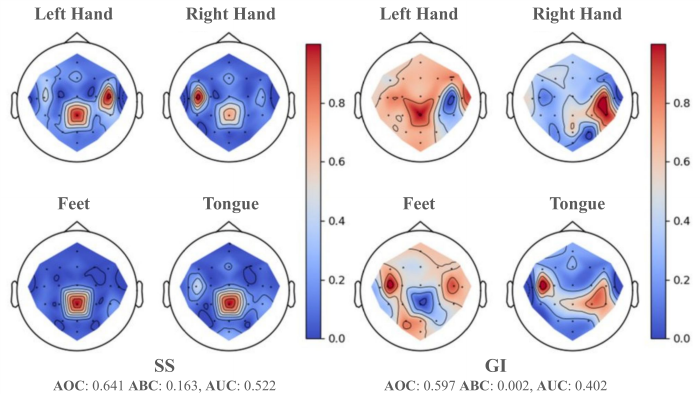}
\caption{
EEG spatial examples on SMR using EEGNet. Saliency values are normalized to $[0,1]$. The most and least faithful methods selected by AIM are compared.
}
\label{fig:qual_eeg}
\end{minipage}

\end{figure}

\paragraph{Audio.}
Figure~\ref{fig:qual_audio} shows that SGA produces clearer frequency-selective saliency profiles than signed gradient-based methods. This supports the quantitative observation that unsigned saliency magnitude can better capture model sensitivity in spectrogram representations, where positive and negative saliency values may cancel during aggregation. Interestingly, GD and SG, which are selected as the most faithful methods by mdROAD and Zeroing, are ranked among the least faithful methods by AIM. This discrepancy suggests that interpolation-based and zero-masking evaluations may favor signed gradient-based explanations in this setting, whereas AIM identifies magnitude-based sensitivity as more consistent with model behavior.
\paragraph{EEG.}
Figure~\ref{fig:qual_eeg} shows a representative SMR spatial example using EEGNet. SmoothGrad Squared captures contralateral motor-cortex patterns for left- and right-hand imagery and responses near the longitudinal fissure for feet imagery, consistent with known sensorimotor rhythms \cite{pfurtscheller2006mu}. This provides qualitative support for the AIM-selected explanation in this setting.

\subsection{Signed and Unsigned Attributions in Signal Domains}

We further examine signed and unsigned attributions in audio and EEG. Signed attributions preserve the direction of contribution, whereas unsigned attributions emphasize magnitude regardless of direction. Treating the two forms as interchangeable can affect ranking-based faithfulness evaluation.

\begin{figure}[hbtp]
\centering
\includegraphics[width=0.9\linewidth]{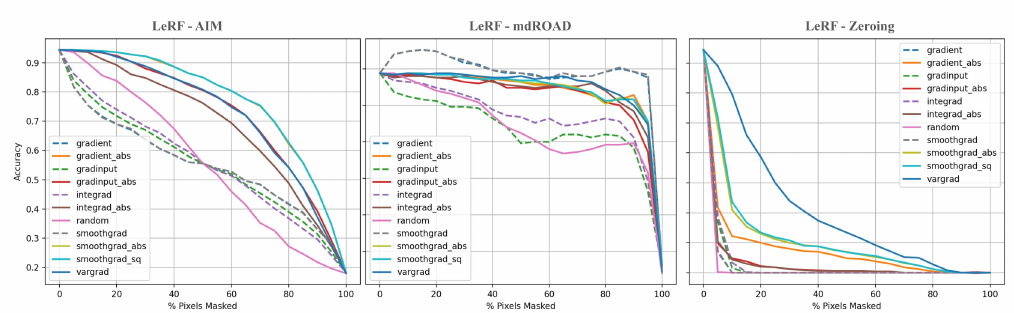}
\caption{
LeRF curves for all explanation methods on MSOS using CNN14. Left: AIM; middle: interpolation-based masking; right: zero masking. Sharp early degradation under LeRF suggests that features assigned low importance may still contain predictive information or that the masking procedure introduces artifacts.
}
\label{fig:msos_cnn14_lerf}
\end{figure}

In temporal-domain settings, including audio waveform and EEG signals, some signed methods show sharp early degradation under LeRF (Figure~\ref{fig:msos_cnn14_lerf}). This suggests that features assigned low signed importance may still contain predictive information, especially when negative attributions are treated as less important. In these cases, attribution magnitude may better capture model sensitivity than attribution sign alone \cite{jang2025timing}.

\begin{figure}[hbtp]
\centering
\includegraphics[width=0.7\linewidth]{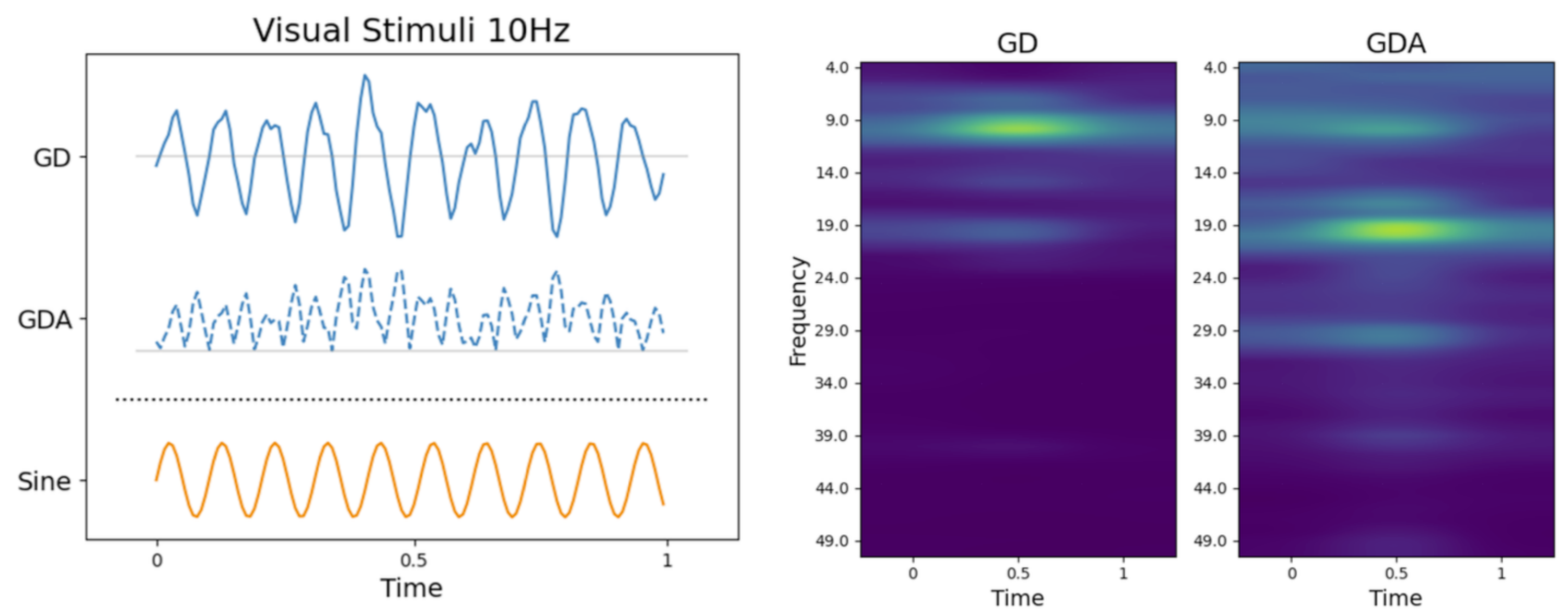}
\caption{
Illustrative effect of absolute-value transformation on a sinusoidal saliency signal. (a) Temporal saliency of Gradient (GD) and its absolute-value variant (GDA) compared with a reference 10-Hz sinusoid. Taking the absolute value halves the apparent period. (b) Time--frequency representation showing that GDA introduces a dominant component near 20 Hz, illustrating how unsigned transformations can distort spectral interpretation.
}
\label{fig:freq_distortion}
\end{figure}

The spectral domain shows a different issue. Taking absolute values can alter the apparent frequency structure of signal-like saliency maps. Figure~\ref{fig:freq_distortion} illustrates this effect using a sinusoidal example: the absolute operation halves the apparent period and introduces a higher harmonic. This does not imply that every unsigned saliency map undergoes the same distortion, but it highlights that unsigned transformations should be interpreted carefully in spectral-domain analysis. Additional frequency-correction analysis is provided in Appendix~\ref{app:frequency_correction}.

Overall, signed and unsigned variants should be selected with respect to both modality and feature domain. In temporal signals, unsigned magnitude may reduce cancellation between positive and negative relevance. In spectral analysis, nonlinear transformations such as taking absolute values may complicate frequency-domain interpretation.

\subsection{Limitations}

Our validation covers image, audio, and EEG tasks, but remains limited in dataset scale, model diversity, and modality coverage. The selected datasets and architectures represent common settings, but do not fully capture the variability of foundation models, transformer-based architectures, large-scale medical data, language models, or other structured signals. The relative performance of AIM, zero masking, and interpolation-based masking may also depend on dataset characteristics, feature granularity, and model robustness. Future work should extend the evaluation to larger datasets, broader model families, additional modalities, and controlled synthetic settings with known ground-truth relevant features.

\section{Conclusion}

We presented AIM, a saliency-guided adversarial information masking framework for evaluating post-hoc saliency maps and masking-operator reliability. By selectively replacing saliency-defined feature subsets with values from a PGD-generated adversarial counterpart, AIM provides an alternative to zero masking and hand-crafted interpolation within a shared MoRF/LeRF protocol. Across image, audio, and EEG tasks, the results suggest that masking-operator choice can substantially affect faithfulness conclusions, and that AIM can reduce some forms of random-attribution bias and ranking instability under the evaluated settings. The signed-versus-unsigned analysis further shows that attribution sign and magnitude should be interpreted with respect to modality and feature domain. 
% Future work may investigate alternative adversarial objectives, stronger distributional constraints, perturbation-matched controls, synthetic ground-truth benchmarks, and applications to language, graph, and multimodal models.

\clearpage
\bibliographystyle{unsrtnat}
\bibliography{references}

@inproceedings{zeiler2014visualizing,
  title={Visualizing and understanding convolutional networks},
  author={Zeiler, Matthew D and Fergus, Rob},
  booktitle={European Conference on Computer Vision (ECCV)},
  pages={818--833},
  year={2014},
  organization={Springer}
}

@article{samek2016evaluating,
  title={Evaluating the visualization of what a deep neural network has learned},
  author={Samek, Wojciech and Binder, Alexander and Montavon, Gregoire and Lapuschkin, Sebastian and Muller, Klaus-Robert},
  journal={IEEE Transactions on Neural Networks and Learning Systems},
  volume={28},
  number={11},
  pages={2660--2673},
  year={2016}
}

@inproceedings{lundberg2017unified,
  title={A unified approach to interpreting model predictions},
  author={Lundberg, Scott M and Lee, Su-In},
  booktitle={Advances in Neural Information Processing Systems},
  volume={30},
  year={2017}
}

@article{ancona2017towards,
  title={Towards better understanding of gradient-based attribution methods for deep neural networks},
  author={Ancona, Marco and Ceolini, Enea and Oztireli, Cengiz and Gross, Markus},
  journal={arXiv preprint arXiv:1711.06104},
  year={2017}
}

@article{goodman2017european,
  title={European union regulations on algorithmic decision-making and a ``right to explanation''},
  author={Goodman, Bryce and Flaxman, Seth},
  journal={AI Magazine},
  volume={38},
  number={3},
  pages={50--57},
  year={2017}
}

@inproceedings{cadamuro2016assessing,
  title={Debugging machine learning models},
  author={Cadamuro, Gabriel and Gilad-Bachrach, Ran and Zhu, Xiaojin},
  booktitle={ICML Workshop on Reliable Machine Learning in the Wild},
  volume={103},
  year={2016}
}

@article{adebayo2020sanity,
  title={Debugging tests for model explanations},
  author={Adebayo, Julius and Muelly, Michael and Liccardi, Ilaria and Kim, Been},
  journal={arXiv preprint arXiv:2011.05429},
  year={2020}
}

@inproceedings{shrikumar2017learning,
  title={Learning important features through propagating activation differences},
  author={Shrikumar, Avanti and Greenside, Peyton and Kundaje, Anshul},
  booktitle={International Conference on Machine Learning},
  pages={3145--3153},
  year={2017},
  organization={PMLR}
}

@inproceedings{adebayo2022post,
  title={Post hoc explanations may be ineffective for detecting unknown spurious correlation},
  author={Adebayo, Julius and Muelly, Michael and Abelson, Harold and Kim, Been},
  booktitle={International Conference on Learning Representations},
  year={2022}
}

@article{roy2019deep,
  title={Deep learning-based electroencephalography analysis: a systematic review},
  author={Roy, Yannick and Banville, Hubert and Albuquerque, Isabela and Gramfort, Alexandre and Falk, Tiago H and Faubert, Jocelyn},
  journal={Journal of Neural Engineering},
  volume={16},
  number={5},
  pages={051001},
  year={2019}
}

@article{tjoa2020survey,
  title={A survey on explainable artificial intelligence (XAI): Toward medical XAI},
  author={Tjoa, Erico and Guan, Cuntai},
  journal={IEEE Transactions on Neural Networks and Learning Systems},
  volume={32},
  number={11},
  pages={4793--4813},
  year={2020}
}

@inproceedings{pan2022explaining,
  title={MATT: A manifold attention network for EEG decoding},
  author={Pan, Yue-Ting and Chou, Jing-Lun and Wei, Chun-Shu},
  booktitle={Advances in Neural Information Processing Systems},
  volume={35},
  pages={31116--31129},
  year={2022}
}

@article{bilodeau2024review,
  title={Impossibility theorems for feature attribution},
  author={Bilodeau, Blair and Jaques, Natasha and Koh, Pang Wei and Kim, Been},
  journal={Proceedings of the National Academy of Sciences},
  volume={121},
  number={2},
  pages={e2304406120},
  year={2024}
}

@inproceedings{shah2021hiding,
  title={Do input gradients highlight discriminative features?},
  author={Shah, Harshay and Jain, Prateek and Netrapalli, Praneeth},
  booktitle={Advances in Neural Information Processing Systems},
  volume={34},
  pages={2046--2059},
  year={2021}
}

@inproceedings{yeh2019fidelity,
  title={On the (in)fidelity and sensitivity of explanations},
  author={Yeh, Chih-Kuan and Hsieh, Cheng-Yu and Suggala, Arun and Inouye, David I and Ravikumar, Pradeep K},
  booktitle={Advances in Neural Information Processing Systems},
  volume={32},
  year={2019}
}

@article{hsieh2020evaluations,
  title={Evaluations and methods for explanation through robustness analysis},
  author={Hsieh, Cheng-Yu and Yeh, Chih-Kuan and Liu, Xuanqing and Ravikumar, Pradeep and Kim, Seungyeon and Kumar, Sanjiv and Hsieh, Cho-Jui},
  journal={arXiv preprint arXiv:2006.00442},
  year={2020}
}

@inproceedings{sundararajan2017axiomatic,
  title={Axiomatic attribution for deep networks},
  author={Sundararajan, Mukund and Taly, Ankur and Yan, Qiqi},
  booktitle={International Conference on Machine Learning},
  pages={3319--3328},
  year={2017},
  organization={PMLR}
}

@incollection{kindermans2019reliability,
  title={The (un)reliability of saliency methods},
  author={Kindermans, Pieter-Jan and Hooker, Sara and Adebayo, Julius and Alber, Maximilian and Sch{\"u}tt, Kristof T and D{\"a}hne, Sven and Erhan, Dumitru and Kim, Been},
  booktitle={Explainable AI: Interpreting, Explaining and Visualizing Deep Learning},
  pages={267--280},
  year={2019},
  publisher={Springer}
}

@inproceedings{adebayo2018sanity,
  title={Sanity checks for saliency maps},
  author={Adebayo, Julius and Gilmer, Justin and Muelly, Michael and Goodfellow, Ian and Hardt, Moritz and Kim, Been},
  booktitle={Advances in Neural Information Processing Systems},
  volume={31},
  year={2018}
}

@article{ravindran2023empirical,
  title={An empirical comparison of deep learning explainability approaches for EEG using simulated ground truth},
  author={Ravindran, Akshay Sujatha and Contreras-Vidal, Jose},
  journal={Scientific Reports},
  volume={13},
  number={1},
  pages={17709},
  year={2023}
}

@inproceedings{tomsett2020sanity,
  title={Sanity checks for saliency metrics},
  author={Tomsett, Richard and Harborne, Dan and Chakraborty, Supriyo and Gurram, Prudhvi and Preece, Alun},
  booktitle={Proceedings of the AAAI Conference on Artificial Intelligence},
  volume={34},
  pages={6021--6029},
  year={2020}
}

@article{brocki2022fidelity,
  title={Fidelity of interpretability methods and perturbation artifacts in neural networks},
  author={Brocki, Lennart and Chung, Neo Christopher},
  journal={arXiv preprint arXiv:2203.02928},
  year={2022}
}

@article{cui2023towards,
  title={Towards best practice of interpreting deep learning models for EEG-based brain computer interfaces},
  author={Cui, Jian and Yuan, Liqiang and Wang, Zhaoxiang and Li, Ruilin and Jiang, Tianzi},
  journal={Frontiers in Computational Neuroscience},
  volume={17},
  pages={1232925},
  year={2023}
}

@article{hooker2018evaluating,
  title={Evaluating feature importance estimates},
  author={Hooker, Sara and Erhan, Dumitru and Kindermans, Pieter-Jan and Kim, Been},
  journal={arXiv preprint arXiv:1806.10758},
  year={2018}
}

@article{rong2022consistent,
  title={A consistent and efficient evaluation strategy for attribution methods},
  author={Rong, Yao and Leemann, Tobias and Borisov, Vadim and Kasneci, Gjergji and Kasneci, Enkelejda},
  journal={arXiv preprint arXiv:2202.00449},
  year={2022}
}

@article{torres2023evaluation,
  title={Evaluation of interpretability for deep learning algorithms in EEG emotion recognition: A case study in autism},
  author={Torres, Juan Manuel Mayor and Medina-DeVilliers, Sara and Clarkson, Tessa and Lerner, Matthew D and Riccardi, Giuseppe},
  journal={Artificial Intelligence in Medicine},
  volume={143},
  pages={102545},
  year={2023}
}

@incollection{park2023geometric,
  title={Geometric remove-and-retrain (GOAR): Coordinate-invariant explainable AI assessment},
  author={Park, Yong-Hyun and Seo, Junghoon and Park, Bomseok and Lee, Seongsu and Jo, Junghyo},
  booktitle={XAI in Action: Past, Present, and Future Applications},
  year={2023},
  publisher={Springer}
}

@article{turbe2023evaluation,
  title={Evaluation of post-hoc interpretability methods in time-series classification},
  author={Turb{\'e}, Hugues and Bjelogrlic, Mina and Lovis, Christian and Mengaldo, Gianmarco},
  journal={Nature Machine Intelligence},
  volume={5},
  number={3},
  pages={250--260},
  year={2023}
}

@inproceedings{fang2024evaluating,
  title={Evaluating post-hoc explanations for graph neural networks via robustness analysis},
  author={Fang, Junfeng and Liu, Wei and Gao, Yuan and Liu, Zemin and Zhang, An and Wang, Xiang and He, Xiangnan},
  booktitle={Advances in Neural Information Processing Systems},
  volume={36},
  year={2024}
}

@inproceedings{dabkowski2017real,
  title={Real time image saliency for black box classifiers},
  author={Dabkowski, Piotr and Gal, Yarin},
  booktitle={Advances in Neural Information Processing Systems},
  volume={30},
  year={2017}
}

@article{goodfellow2014explaining,
  title={Explaining and harnessing adversarial examples},
  author={Goodfellow, Ian J and Shlens, Jonathon and Szegedy, Christian},
  journal={arXiv preprint arXiv:1412.6572},
  year={2014}
}

@article{zhang2019vulnerability,
  title={On the vulnerability of CNN classifiers in EEG-based BCIs},
  author={Zhang, Xiao and Wu, Dongrui},
  journal={IEEE Transactions on Neural Systems and Rehabilitation Engineering},
  volume={27},
  number={5},
  pages={814--825},
  year={2019}
}

@article{nieradzik2025reliable,
  title={Reliable Evaluation of Attribution Maps in CNNs: A Perturbation-Based Approach},
  author={Nieradzik, Laura and Stephani, Heike and Keuper, Jan},
  journal={International Journal of Computer Vision},
  volume={133},
  pages={2392--2409},
  year={2025},
  doi={10.1007/s11263-024-02282-6}
}

@article{becker2024audiomnist,
  title={AudioMNIST: Exploring Explainable Artificial Intelligence for audio analysis on a simple benchmark},
  author={Becker, S{\"o}ren and Vielhaben, Johanna and Ackermann, Marcel and M{\"u}ller, Klaus-Robert and Lapuschkin, Sebastian and Samek, Wojciech},
  journal={Journal of the Franklin Institute},
  volume={361},
  number={1},
  pages={418--428},
  year={2024},
  doi={10.1016/j.jfranklin.2023.11.038}
}

@article{apicella2022toward,
  title={Toward the application of XAI methods in EEG-based systems},
  author={Apicella, Andrea and Isgr{\`o}, Francesco and Pollastro, Andrea and Prevete, Roberto},
  journal={arXiv preprint arXiv:2210.06554},
  year={2022}
}

@article{singh2021towards,
  title={Towards bridging the gap between computational intelligence and neuroscience in brain-computer interfaces},
  author={Singh, Avinash Kumar and Sahonero-Alvarez, Guillermo and Mahmud, Mufti and Bianchi, Luigi},
  journal={Frontiers in Neuroinformatics},
  volume={15},
  pages={699840},
  year={2021}
}

@article{rajpura2024explainable,
  title={Explainable artificial intelligence approaches for brain-computer interfaces: A review and design space},
  author={Rajpura, Param and Cecotti, Hubert and Meena, Yogesh Kumar},
  journal={Journal of Neural Engineering},
  year={2024}
}

@article{friedrich2020mfbb,
  title={Multipoint fractional Brownian bridge: construction and applications},
  author={Friedrich, Tobias and et al.},
  journal={Stochastic Processes and their Applications},
  year={2020}
}

@article{donoghue2020parameterizing,
  title={Parameterizing neural power spectra into periodic and aperiodic components},
  author={Donoghue, Thomas and Haller, Matthijs and Peterson, Erik J and Varma, Paroma and Sebastian, Padraig and Gao, Ruijie and Noto, Takashi and Lara, Antonio H and Wallis, Jonathan D and Knight, Robert T and others},
  journal={Nature Neuroscience},
  volume={23},
  number={12},
  pages={1655--1665},
  year={2020}
}

@article{he2014scale,
  title={Scale-free brain activity: past, present, and future},
  author={He, Biyu J.},
  journal={Trends in Cognitive Sciences},
  volume={18},
  number={9},
  pages={480--487},
  year={2014}
}

@book{beran2013long,
  title={Long-memory processes: probabilistic properties and statistical methods},
  author={Beran, Jan},
  year={2013},
  publisher={Springer}
}

@article{banville2022robust,
  title={Robust learning from corrupted EEG with self-supervised learning},
  author={Banville, Hubert and Chehab, Omar and Hyvarinen, Aapo and Engemann, Denis-Alexandre and Gramfort, Alexandre},
  journal={NeuroImage},
  volume={251},
  pages={118994},
  year={2022}
}

@article{mandelbrot1968fractional,
  title={Fractional Brownian motions, fractional noises and applications},
  author={Mandelbrot, Benoit B. and Van Ness, John W.},
  journal={SIAM Review},
  volume={10},
  number={4},
  pages={422--437},
  year={1968}
}

@article{dieker2004simulation,
  title={Simulation of fractional Brownian motion},
  author={Dieker, Ton},
  journal={Master’s thesis, University of Twente},
  year={2004}
}

@article{davies1987tests,
  title={Tests for Hurst effect},
  author={Davies, Robert B. and Harte, D. S.},
  journal={Biometrika},
  volume={74},
  number={1},
  pages={95--101},
  year={1987}
}

@article{madry2017towards,
  title={Towards deep learning models resistant to adversarial attacks},
  author={Madry, Aleksander and Makelov, Aleksandar and Schmidt, Ludwig and Tsipras, Dimitris and Vladu, Adrian},
  journal={arXiv preprint arXiv:1706.06083},
  year={2017}
}

@article{leske2019reducing,
  title={Reducing power line noise in EEG and MEG data via spectrum interpolation},
  author={Leske, Stephan and Dalal, Sarang S.},
  journal={NeuroImage},
  volume={189},
  pages={763--776},
  year={2019}
}

@article{kannathal2005analysis,
  title={Analysis of EEG signals using fractal dimension},
  author={Kannathal, N. and et al.},
  journal={Biomedical Signal Processing and Control},
  year={2005}
}

@inproceedings{kroos2019msos,
  title={Generalisation in environmental sound classification: The Making Sense of Sounds dataset},
  author={Kroos, C. and Bones, O. and Cao, Y. and Harris, L. and Jackson, P. J. and Davies, W. J. and Wang, W. and Cox, T. J. and Plumbley, M. D.},
  booktitle={ICASSP},
  pages={8082--8086},
  year={2019}
}

@inproceedings{piczak2015dataset,
  title={ESC: Dataset for Environmental Sound Classification},
  author={Piczak, Karol J.},
  booktitle={ACM Multimedia},
  year={2015}
}

@inproceedings{deng2009imagenet,
  title={ImageNet: A large-scale hierarchical image database},
  author={Deng, Jia and Dong, Wei and Socher, Richard and Li, Li-Jia and Li, Kai and Fei-Fei, Li},
  booktitle={Proceedings of the IEEE Conference on Computer Vision and Pattern Recognition (CVPR)},
  year={2009}
}

@inproceedings{parkhi2012cats,
  title={Cats and dogs},
  author={Parkhi, Omkar M. and Vedaldi, Andrea and Zisserman, Andrew},
  booktitle={CVPR},
  year={2012}
}

@misc{bhuvaji2020brain,
  title={Brain Tumor Classification (MRI)},
  author={Bhuvaji, S. and Kadam, A. and Bhumkar, P. and Dedge, S.},
  year={2020},
  howpublished={Kaggle}
}

@article{lawhern2018eegnet,
  title={EEGNet: A compact convolutional neural network for EEG-based BCIs},
  author={Lawhern, Vernon J. and et al.},
  journal={Journal of Neural Engineering},
  year={2018}
}

@article{cui2022interpretable,
  title={Interpretable CNN for EEG},
  author={Cui, Zhiyuan and et al.},
  journal={IEEE TBME},
  year={2022}
}

@article{brunner2008bci,
  title={BCI Competition 2008–Graz data set A},
  author={Brunner, Clemens and Leeb, Robert and M{\"u}ller-Putz, Gernot R and Schl{\"o}gl, Alois and Pfurtscheller, Gert},
  journal={Institute for Knowledge Discovery (Laboratory of Brain-Computer Interfaces), Graz University of Technology},
  year={2008}
}

@misc{mattout2014bci,
  title={BCI Challenge: Event-Related Negativity Dataset},
  author={Mattout, J{\'e}r{\'e}mie and Kan, K},
  year={2014},
  note={Kaggle dataset},
  url={https://www.kaggle.com/c/inria-bci-challenge}
}

@article{martinez2007mamem,
  title={A multi-modal dataset for steady-state visual evoked potential-based brain-computer interfaces},
  author={Martinez-Cagigal, Victor and Santamaria-Vazquez, Enrique and Hornero, Roberto},
  journal={MAMEM SSVEP Dataset},
  year={2007}
}

@inproceedings{simonyan2013deep,
  title={Deep Inside Convolutional Networks: Visualising Image Classification Models and Saliency Maps},
  author={Simonyan, Karen and Vedaldi, Andrea and Zisserman, Andrew},
  booktitle={arXiv preprint arXiv:1312.6034},
  year={2013}
}

@inproceedings{smilkov2017smoothgrad,
  title={SmoothGrad: removing noise by adding noise},
  author={Smilkov, Daniel and Thorat, Nikhil and Kim, Been and Vi{\'e}gas, Fernanda and Wattenberg, Martin},
  booktitle={arXiv preprint arXiv:1706.03825},
  year={2017}
}

@inproceedings{selvaraju2017grad,
  title={Grad-CAM: Visual Explanations from Deep Networks via Gradient-based Localization},
  author={Selvaraju, Ramprasaath R. and Cogswell, Michael and Das, Abhishek and Vedantam, Ramakrishna and Parikh, Devi and Batra, Dhruv},
  booktitle={ICCV},
  year={2017}
}

@inproceedings{hooker2019benchmark,
  title={A benchmark for interpretability methods in deep neural networks},
  author={Hooker, Sara and Erhan, Dumitru and Kindermans, Pieter-Jan and Kim, Been},
  booktitle={NeurIPS},
  year={2019}
}

@article{bach2015pixel,
  title={On pixel-wise explanations for non-linear classifier decisions by layer-wise relevance propagation},
  author={Bach, Sebastian and others},
  journal={PLOS ONE},
  year={2015}
}

@inproceedings{ancona2018towards,
  title={Towards better understanding of gradient-based attribution methods},
  author={Ancona, Marco and others},
  booktitle={ICLR},
  year={2018}
}

@inproceedings{chattopadhyay2018grad,
  title={Grad-CAM++: Improved visual explanations for deep convolutional networks},
  author={Chattopadhyay, Aditya and others},
  booktitle={WACV},
  year={2018}
}

@article{apicella2022towards,
  title={Towards robust evaluation of explainable artificial intelligence methods},
  author={Apicella, Andrea and others},
  journal={Pattern Recognition Letters},
  year={2022}
}

@inproceedings{brocki2022evaluation,
  title={On the evaluation of saliency methods},
  author={Brocki, Lukasz and Chung, Ngan},
  booktitle={ICLR Workshop},
  year={2022}
}

@article{cui2023explaining,
  title={Explaining deep learning models: A comprehensive survey},
  author={Cui, Zhiyong and others},
  journal={IEEE Transactions},
  year={2023}
}

@inproceedings{wei2019eeg,
  title={Spatial Component-wise Convolutional Network (SCCNet) for Motor-Imagery EEG Classification},
  author={Wei, Chun-Shu and Koike-Akino, Toshiaki and Wang, Ye},
  booktitle={Proceedings of the 9th International IEEE/EMBS Conference on Neural Engineering (NER)},
  pages={328--331},
  year={2019},
  organization={IEEE}
}

@inproceedings{
jang2025timing,
title={{TIMING}: Temporality-Aware Integrated Gradients for Time Series Explanation},
author={Hyeongwon Jang and Changhun Kim and Eunho Yang},
booktitle={Forty-second International Conference on Machine Learning},
year={2025},
url={https://openreview.net/forum?id=qOgKMqv9T7}
}

@INPROCEEDINGS{9150840,
  author={Wang, Haofan and Wang, Zifan and Du, Mengnan and Yang, Fan and Zhang, Zijian and Ding, Sirui and Mardziel, Piotr and Hu, Xia},
  booktitle={2020 IEEE/CVF Conference on Computer Vision and Pattern Recognition Workshops (CVPRW)}, 
  title={Score-CAM: Score-Weighted Visual Explanations for Convolutional Neural Networks}, 
  year={2020},
  volume={},
  number={},
  pages={111-119},
  doi={10.1109/CVPRW50498.2020.00020}}

@article{omeiza2019smooth,
  title={Smooth grad-cam++: An enhanced inference level visualization technique for deep convolutional neural network models},
  author={Omeiza, Daniel and Speakman, Skyler and Cintas, Celia and Weldermariam, Komminist},
  journal={arXiv preprint arXiv:1908.01224},
  year={2019}
}

@article{pfurtscheller2006mu,
  title={Mu rhythm (de)synchronization and EEG single-trial classification of different motor imagery tasks},
  author={Pfurtscheller, Gert and Brunner, Clemens and Schl{\"o}gl, Alois and da Silva, F. H. Lopes},
  journal={NeuroImage},
  volume={31},
  number={1},
  pages={153--159},
  year={2006},
  publisher={Elsevier}
}

@inproceedings{zheng2025ffidelity,
  title={F-Fidelity: A Robust Framework for Faithfulness Evaluation of Explainable AI},
  author={Zheng, Xu and Shirani, Farhad and Chen, Zhuomin and Lin, Chaohao and Cheng, Wei and Guo, Wenbo and Luo, Dongsheng},
  booktitle={International Conference on Learning Representations},
  year={2025},
  note={Poster}
}

@article{skliarov2025comparative,
  title={A Comparative Evaluation of Explainability Techniques for Image Data},
  author={Skliarov, Mykyta and El Shawi, Radwa and Dhaoui, Chedia and others},
  journal={Scientific Reports},
  volume={15},
  pages={41898},
  year={2025},
  doi={10.1038/s41598-025-25839-y}
}

@misc{gomez2022metrics,
  title={Metrics for Saliency Map Evaluation of Deep Learning Explanation Methods},
  author={Gomez, Thibault and Fr{\'e}our, Thomas and Mouch{\`e}re, Harold},
  year={2022},
  eprint={2201.13291},
  archivePrefix={arXiv},
  primaryClass={cs.CV}
}

\appendix

\section{Appendix}
\subsection{Compute Resources}
\label{app:compute_resources}

All experiments were conducted on GPU-based workstations. The main experiments were run using NVIDIA RTX 4090 GPUs with CUDA acceleration. For each dataset--model--masking configuration, we generated post-hoc saliency maps, constructed MoRF and LeRF masking orders, and evaluated model performance under AIM, mdROAD, and Zeroing. AIM additionally required one PGD-generated adversarial counterpart for each evaluated input sample, which increased the evaluation cost relative to Zeroing and mdROAD.

For EEG experiments, each decoder was trained in a subject-specific manner and repeated across five random seeds. Training a single EEG decoder typically required less than one hour, while saliency generation and masking-based evaluation varied depending on the number of subjects, feature domains, and explanation methods. Audio and image experiments were more computationally demanding because they involved larger input sizes and pretrained CNN architectures. Across modalities, the most expensive step was the repeated evaluation of masked inputs over multiple masking ratios and explanation methods.

In total, the experiments covered three modalities, nine datasets, multiple neural architectures, three masking frameworks, and multiple post-hoc explanation methods. Quantitative results are reported as mean $\pm$ standard deviation across datasets, models, subjects, or repeated runs depending on the modality-specific experimental design. The computational cost is moderate and comparable to standard post-hoc explanation benchmarking pipelines, with the additional overhead of PGD-based adversarial replacement in AIM.

\subsection{Adversarial Perturbation Details for AIM}

\paragraph{Attack method for AIM framework.}
We adopt Projected Gradient Descent (PGD) to construct adversarial perturbations. Starting from the original input $x$, a small perturbation is first added, followed by iterative gradient updates that aim to maximize the classification loss while constraining the perturbation magnitude.

At iteration $t$, the update is defined as:
\[
x^{\mathrm{adv}}_{t+1} = \Pi_{\mathcal{B}_\epsilon(x)} \left( x^{\mathrm{adv}}_t + \alpha \cdot \mathrm{sign} \left( \nabla_x \mathcal{L}(f(x^{\mathrm{adv}}_t), y) \right) \right),
\]
where $\mathcal{L}$ denotes the loss function, $f(\cdot)$ is the model, $\alpha$ is the step size, and $\Pi_{\mathcal{B}_\epsilon(x)}$ projects the perturbed sample back to an $\ell_2$ or $\ell_\infty$ ball of radius $\epsilon$ centered at $x$.

The hyperparameters are selected empirically: we set the step size $\alpha = 2$ and the number of iterations to 10. The perturbation radius $\epsilon$ is chosen such that, under full masking (i.e., when all features are replaced by adversarial counterparts), the model performance is reduced to approximately chance level. This ensures that the perturbation is sufficiently strong to eliminate predictive information while remaining within a controlled magnitude.

\subsection{FEATURE IMPUTATION IN mdROAD}

\paragraph{Spatial domain (EEG channels / spectrogram / image).}
For spatial features, we select a subset of channels $\Phi_c$ corresponding to the top-$k\%$ (MoRF) or bottom-$k\%$ (LeRF) importance scores in $S_c$. The imputation is performed using weighted interpolation from neighboring features to preserve local spatial structure.

For EEG, the interpolation is defined based on the electrode montage. Each target channel is reconstructed using its neighboring channels as:
\[
x'_{c \in \Phi_c, t} = \sum_{c' \notin \Phi_c} W_{c,c'} \, x_{c',t} + \epsilon,
\]
where $W$ is a spatial weight matrix and $\epsilon$ is small noise to mitigate information leakage.

Specifically, we adopt a Laplacian-style spatial filter, where each electrode is connected to a set of neighboring electrodes consisting of four direct neighbors and four diagonal (indirect) neighbors. The weights $w_d$ = 1/6 and $w_{id}$ = 1/12 are assigned for direct and indirect neighbors, respectively, ensuring that the total contribution sums to one. This formulation is commonly used for channel-wise interpolation in EEG analysis  \cite{banville2022robust}.

Although the selected channels $\Phi_c$ are not necessarily contiguous, they often form connected regions in practice. In such cases, the imputation can be solved jointly as a sparse linear system to ensure consistency across neighboring interpolations.

This spatial interpolation strategy naturally extends to other modalities. For spectrograms and images, neighboring pixels or time-frequency bins are used to construct $W$, enabling a unified in-distribution imputation framework across domains.

\paragraph{Spectral domain (EEG frequency).}
For spectral features, we select a contiguous frequency band $\Phi_f$ whose total contribution accounts for $k\%$ of the sample power and is ranked as the most (MoRF) or least (LeRF) important according to $S_f$. The bandwidth is determined via exhaustive search over all possible contiguous intervals (details in Appendix B.4), allowing the selected region to adapt to different samples and configurations.

To preserve the intrinsic structure of the EEG spectrum, we design a spectral imputation strategy inspired by noisy linear interpolation  \cite{rong2022consistent} and spectrum interpolation methods used in signal denoising  \cite{leske2019reducing}. Empirical studies suggest that neural power spectra exhibit a $1/f$-like aperiodic structure  \cite{he2014scale, donoghue2020parameterizing}, which reflects scale-free dynamics arising from correlated activity across frequencies.

To maintain this structure, we fit a real-valued polynomial of degree three to the observed power spectrum outside the masked region:
\[
P(f) = \sum_{i=0}^{3} a_i f^{-i},
\]
where the coefficients $\{a_i\}$ are estimated from $X_{c,f \notin \Phi_f}$. The imputed spectrum is then defined as:
\[
X'_{c,f \in \Phi_f} = P(f),
\]
while the remaining frequencies are kept unchanged. The time-domain signal is reconstructed via inverse Fourier transform:
\[
x'_{c,t} = \mathcal{F}_c^{-1}(X'_{c,f}).
\]

Notably, since phase information cannot be reliably inferred from neighboring frequencies and is weakly structured across frequency bins, we only modify the amplitude of the spectrum while preserving the original phase. This design intentionally disrupts the linear relationship between frequency components while maintaining realistic spectral statistics, resulting in an in-distribution yet semantically corrupted signal.

\paragraph{Temporal domain (EEG / waveform audio).}
For temporal features, we select a contiguous time interval $\Phi_t$ whose length corresponds to $k\%$ of the signal duration and whose cumulative importance is ranked as the most (MoRF) or least (LeRF) according to $S_t$.

To generate in-distribution imputations that preserve temporal dynamics, we adopt the Multipoint Fractional Brownian Bridge (MFBB)  \cite{friedrich2020mfbb}. MFBB is a self-similar stochastic process designed for interpolating sparsely observed time series, parameterized by a Hurst index $H$ and conditioned on a set of observed anchor points $\{(t_i, G_i)\}$. This formulation is particularly suitable for modeling natural signals with long-range dependencies.

The Hurst index $H$ controls the temporal correlation structure of the generated process  \cite{beran2013long, kannathal2005analysis}. In particular, $H < 0.5$ corresponds to anti-persistent (mean-reverting) behavior, which discourages smooth extrapolation and reduces the risk of preserving class-discriminative patterns  \cite{mandelbrot1968fractional}. In our implementation, we set $H = 10^{-5}$ to enforce strong anti-persistence.

For each masked interval $\Phi_t$, we define three anchor points located at the beginning, center, and end of the interval. The corresponding values $\{G_i\}$ are taken from the original signal, ensuring minimal structural consistency while limiting information leakage.

Let $B(t)$ denote a fractional Brownian motion (FBM) process with covariance:
\[
<B(t_1), B(t_2)> = \frac{1}{2} \left( |t_1|^{2H} + |t_2|^{2H} - |t_1 - t_2|^{2H} \right),
\]
which can be efficiently sampled using the Davies–Harte method  \cite{davies1987tests, dieker2004simulation}. The MFBB constructs a conditional process that interpolates the anchor points while preserving the covariance structure. The imputed signal is given by:
\[
x'_{c,t \in \Phi_t} = B(t) - \bigl(B(t_i) - G_i\bigr)\sigma_{ij}^{-1} \langle B(t), B(t_j) \rangle,
\]
where $\sigma_{ij}$ denotes the covariance matrix between anchor points.

This formulation ensures that the imputed segment remains consistent with the statistical properties of natural time series while disrupting task-relevant temporal patterns. The same approach naturally extends to waveform audio, where preserving temporal continuity and long-range dependency is critical for maintaining perceptual plausibility.

\subsection{Experimental Setup Details}
\label{app:experimental_setup_details}

\paragraph{Image.}
For visual data, we use both natural and medical image classification tasks to cover diverse semantic and structural characteristics. We evaluate three representative CNN architectures: ResNet-50, EfficientNet-B0, and RepVGG-B0. These models provide different architectural biases, including residual learning, parameter-efficient scaling, and VGG-style re-parameterization. Since image inputs are treated as a spatial domain in our framework, they provide a practical setting for evaluating zero masking, interpolation-based imputation, and adversarial information masking in two-dimensional structured data.

\paragraph{Audio.}
For audio data, we include both speech and environmental sound classification tasks, covering waveform-based and spectrogram-based representations. This allows us to examine whether faithfulness evaluation behaves consistently across different acoustic structures. For waveform inputs, we use AudioNet \cite{becker2024audiomnist} and the pre-trained Res1dNet31 \cite{nieradzik2025reliable}, which directly process temporal signals. For spectrogram inputs, we use AlexNet \cite{becker2024audiomnist} adapted to audio classification and the pre-trained CNN14 \cite{nieradzik2025reliable}, which captures time--frequency representations. These model families provide complementary inductive biases: waveform models emphasize temporal structure, while spectrogram models capture spatial patterns over time--frequency bins.

\paragraph{EEG.}
For neural time-series data, we use three EEG datasets representing distinct BCI paradigms. Compared with image and audio data, EEG introduces additional challenges, including high inter-subject variability, low signal-to-noise ratio, and structured dependencies across spatial, temporal, and spectral domains. We evaluate three CNN-based EEG decoders: EEGNet \cite{lawhern2018eegnet}, SCCNet \cite{wei2019eeg}, and InterpretableCNN \cite{cui2022interpretable}. These models represent compact architectures, spatial-filtering-oriented designs, and interpretability-oriented designs. All EEG models are trained in a subject-specific manner to reduce inter-subject variability and align with standard EEG decoding practice. Detailed preprocessing, training configurations, and hyperparameters are provided in the released implementation.

\subsection{Full Faithfulness Evaluation Results}
\label{app:full_results}

\begin{table}[H]
\centering
\tiny
\setlength{\tabcolsep}{5pt}
\renewcommand{\arraystretch}{0.4} 

\resizebox{\linewidth}{!}{%
\begin{tabular}{l|ccc|ccc|ccc}
\toprule
\textbf{Method} 
& \multicolumn{3}{c|}{AIM}
& \multicolumn{3}{c|}{mdROAD} 
& \multicolumn{3}{c}{Zeroing} \\

& AOC & ABC & AUC & AOC & ABC & AUC & AOC & ABC & AUC \\
\midrule

GC
& .724$\pm$.059 & .533$\pm$.100 & \textbf{.808$\pm$.070}
& \textbf{.800$\pm$.077} & \textbf{.596$\pm$.275} & \textbf{.783$\pm$.232}
& .762$\pm$.099 & \textbf{.555$\pm$.133} & \textbf{.788$\pm$.169} \\

GC++
& .722$\pm$.065 & .531$\pm$.106 & \textbf{.808$\pm$.070}
& .788$\pm$.083 & .567$\pm$.272 & .771$\pm$.228
& .751$\pm$.108 & .531$\pm$.136 & .779$\pm$.168 \\

SC
& .716$\pm$.062 & .522$\pm$.105 & .805$\pm$.073
& .785$\pm$.080 & .569$\pm$.263 & .776$\pm$.224
& .755$\pm$.129 & .535$\pm$.141 & .777$\pm$.184 \\

SGC++
& \textbf{.728$\pm$.068} & \textbf{.538$\pm$.107} & \textbf{.808$\pm$.069}
& .798$\pm$.070 & .581$\pm$.252 & .776$\pm$.220
& .758$\pm$.113 & .536$\pm$.120 & .777$\pm$.165 \\

GD
& .667$\pm$.108 & .362$\pm$.176 & .694$\pm$.101
& .713$\pm$.115 & .426$\pm$.235 & .711$\pm$.160
& .826$\pm$.093 & .314$\pm$.212 & .484$\pm$.211 \\

SG
& .683$\pm$.111 & .404$\pm$.184 & .720$\pm$.102
& .730$\pm$.117 & .458$\pm$.240 & .723$\pm$.162
& .824$\pm$.082 & .348$\pm$.231 & .518$\pm$.218 \\

GI
& .636$\pm$.099 & .328$\pm$.162 & .690$\pm$.100
& .669$\pm$.128 & .363$\pm$.196 & .690$\pm$.134
& .819$\pm$.092 & .330$\pm$.143 & .508$\pm$.166 \\

IG
& .672$\pm$.088 & .396$\pm$.140 & .721$\pm$.091
& .702$\pm$.107 & .430$\pm$.191 & .724$\pm$.144
& \textbf{.843$\pm$.075} & .394$\pm$.164 & .542$\pm$.211 \\

Random
& .527$\pm$.060 & .000$\pm$.004 & .473$\pm$.062
& .535$\pm$.094 & .022$\pm$.033 & .487$\pm$.086
& .763$\pm$.105 & -.001$\pm$.004 & .226$\pm$.121 \\

\bottomrule
\end{tabular}
}
\caption{Image-domain faithfulness scores under AIM, interpolation-based masking, and zero masking. Values report mean $\pm$ standard deviation across image datasets and models. Higher AOC indicates stronger MoRF degradation, higher AUC indicates stronger LeRF preservation, and higher ABC indicates greater MoRF-LeRF separation. Best values in each column are in bold.}
\label{tab:full_image}
\end{table}

\begin{table}[H]
\centering
\tiny
\setlength{\tabcolsep}{5pt}
\renewcommand{\arraystretch}{0.4} 
\resizebox{\textwidth}{!}{%
\begin{tabular}{l|ccc|ccc|ccc}
\toprule
\textbf{Method} & \multicolumn{3}{c|}{AIM} & \multicolumn{3}{c|}{mdROAD} & \multicolumn{3}{c}{Zeroing} \\
& AOC & ABC & AUC & AOC & ABC & AUC & AOC & ABC & AUC \\
\midrule
GD  & .576$\pm$.079 & .007$\pm$.009 & .423$\pm$.084 
    & .649$\pm$.213 & .015$\pm$.022 & .349$\pm$.208 
    & .750$\pm$.129 & .010$\pm$.019 & .250$\pm$.141 \\

GI  & .593$\pm$.061 & .015$\pm$.020 & .420$\pm$.073 
    & .699$\pm$.190 & .101$\pm$.023 & .395$\pm$.196 
    & \textbf{.822$\pm$.108} & .135$\pm$.145 & .310$\pm$.201 \\

VG  & .626$\pm$.077 & .125$\pm$.117 & .493$\pm$.107 
    & .518$\pm$.129 & .023$\pm$.035 & .411$\pm$.185 
    & .588$\pm$.243 & .018$\pm$.030 & .281$\pm$.131 \\

SG  & .589$\pm$.065 & .009$\pm$.006 & .415$\pm$.066 
    & .649$\pm$.207 & .011$\pm$.017 & .352$\pm$.209 
    & .755$\pm$.130 & .012$\pm$.014 & .252$\pm$.142 \\

IG  & .593$\pm$.057 & .022$\pm$.030 & .426$\pm$.076 
    & \textbf{.719$\pm$.165} & .153$\pm$.057 & .427$\pm$.220 
    & \textbf{.822$\pm$.102} & .142$\pm$.132 & .320$\pm$.199 \\

GDA & \textbf{.694$\pm$.066} & \textbf{.261$\pm$.170} 
    & \textbf{.567$\pm$.142} 
    & .637$\pm$.165 & .078$\pm$.058 & .434$\pm$.211 
    & .740$\pm$.105 & .108$\pm$.105 & .363$\pm$.186 \\

GIA & .672$\pm$.062 & .200$\pm$.160 & .527$\pm$.133 
    & .705$\pm$.165 & .292$\pm$.203 & .584$\pm$.150 
    & .814$\pm$.107 & .293$\pm$.286 & .475$\pm$.279 \\

SGA & .677$\pm$.065 & .228$\pm$.166 & .551$\pm$.139 
    & .596$\pm$.153 & .059$\pm$.041 & .447$\pm$.182 
    & .710$\pm$.128 & .085$\pm$.088 & .358$\pm$.188 \\

SS  & .675$\pm$.064 & .222$\pm$.160 & .547$\pm$.136 
    & .594$\pm$.147 & .059$\pm$.035 & .447$\pm$.183 
    & .700$\pm$.138 & .079$\pm$.088 & .353$\pm$.187 \\

IGA & .659$\pm$.059 & .174$\pm$.151 & .515$\pm$.129 
    & .710$\pm$.163 & \textbf{.307$\pm$.219} & \textbf{.592$\pm$.155} 
    & .816$\pm$.107 & \textbf{.302$\pm$.297} & \textbf{.481$\pm$.287} \\

Random  & .570$\pm$.063 & .003$\pm$.004 & .427$\pm$.064 
    & .637$\pm$.216 & .006$\pm$.006 & .363$\pm$.212 
    & .726$\pm$.143 & .003$\pm$.002 & .270$\pm$.135 \\

\bottomrule
\end{tabular}
}
\caption{Audio waveform-domain faithfulness scores for AudioNet and Res1dNet31. Values report mean $\pm$ standard deviation across audio datasets and models. Best values in each column are in bold.}
\label{tab:full_audio_wave}
\end{table}

\begin{table}[H]
\centering
\tiny
\setlength{\tabcolsep}{5pt}
\renewcommand{\arraystretch}{0.4} 
\resizebox{\textwidth}{!}{%
\begin{tabular}{l|ccc|ccc|ccc}
\toprule
\textbf{Method} & \multicolumn{3}{c|}{AIM} & \multicolumn{3}{c|}{mdROAD} & \multicolumn{3}{c}{Zeroing} \\
& AOC & ABC & AUC & AOC & ABC & AUC & AOC & ABC & AUC \\
\midrule
GD  & .594$\pm$.153 & .003$\pm$.004 & .402$\pm$.150 
    & \textbf{.724$\pm$.236} & \textbf{.657$\pm$.299} & .733$\pm$.344 
    & \textbf{.940$\pm$.029} & .267$\pm$.390 & .293$\pm$.351 \\

GI  & .603$\pm$.132 & .024$\pm$.027 & .421$\pm$.149 
    & .656$\pm$.375 & .555$\pm$.450 & .681$\pm$.336 
    & .875$\pm$.151 & .166$\pm$.368 & .208$\pm$.264 \\

VG  & .693$\pm$.099 & .290$\pm$.180 & .569$\pm$.181 
    & .480$\pm$.131 & .300$\pm$.225 & .748$\pm$.141 
    & .842$\pm$.164 & .167$\pm$.197 & .293$\pm$.245 \\

SG  & .594$\pm$.153 & .003$\pm$.004 & .399$\pm$.150 
    & .722$\pm$.241 & .655$\pm$.306 & .734$\pm$.344 
    & \textbf{.940$\pm$.028} & \textbf{.268$\pm$.390} & .291$\pm$.346 \\

IG  & .603$\pm$.128 & .030$\pm$.027 & .426$\pm$.137 
    & .654$\pm$.357 & .566$\pm$.435 & .701$\pm$.339 
    & .873$\pm$.150 & .169$\pm$.367 & .219$\pm$.269 \\

GDA & .770$\pm$.065 & .427$\pm$.127 & \textbf{.657$\pm$.136} 
    & .578$\pm$.171 & .367$\pm$.138 & \textbf{.777$\pm$.084} 
    & .909$\pm$.059 & .228$\pm$.246 & .309$\pm$.299 \\

GIA & .726$\pm$.078 & .330$\pm$.115 & .604$\pm$.140 
    & .560$\pm$.175 & .359$\pm$.164 & .758$\pm$.099 
    & .891$\pm$.089 & .175$\pm$.233 & .263$\pm$.282 \\

SGA & \textbf{.774$\pm$.058} & \textbf{.431$\pm$.125} 
    & \textbf{.657$\pm$.139} 
    & .576$\pm$.158 & .374$\pm$.149 & .769$\pm$.100 
    & .900$\pm$.075 & .241$\pm$.248 & .334$\pm$.314 \\

SS  & .772$\pm$.059 & .426$\pm$.133 & .653$\pm$.143 
    & .558$\pm$.156 & .362$\pm$.157 & .769$\pm$.100 
    & .900$\pm$.075 & .242$\pm$.250 & \textbf{.336$\pm$.316} \\

IGA & .693$\pm$.104 & .252$\pm$.061 & .559$\pm$.112 
    & .512$\pm$.145 & .347$\pm$.132 & .740$\pm$.166 
    & .888$\pm$.109 & .160$\pm$.234 & .249$\pm$.265 \\

Random  & .571$\pm$.111 & .001$\pm$.000 & .429$\pm$.111 
    & .390$\pm$.093 & .088$\pm$.079 & .633$\pm$.146 
    & .846$\pm$.160 & .002$\pm$.004 & .154$\pm$.160 \\
\bottomrule
\end{tabular}
}
\caption{Audio spectrogram-domain faithfulness scores for AlexNet and CNN14. Values report mean $\pm$ standard deviation across audio datasets and models. Best values in each column are in bold.}
\label{tab:full_audio_spec}
\end{table}

% ================= Spatial =================
\begin{table}[H]
\centering
\tiny
\setlength{\tabcolsep}{5pt}
\renewcommand{\arraystretch}{0.4} 
\resizebox{\textwidth}{!}{%

\begin{tabular}{l|ccc|ccc|ccc}
\toprule
& \multicolumn{3}{c|}{AIM}
& \multicolumn{3}{c|}{mdROAD}
& \multicolumn{3}{c}{Zeroing} \\
\textbf{Method}
& AOC & ABC & AUC
& AOC & ABC & AUC
& AOC & ABC & AUC \\
\midrule
GD  & .492$\pm$.143 & .002$\pm$.002 & .508$\pm$.143 & .491$\pm$.181 & .024$\pm$.042 & .463$\pm$.095 & .763$\pm$.151 & .148$\pm$.163 & .287$\pm$.206 \\
GI  & .494$\pm$.151 & .008$\pm$.010 & .513$\pm$.149 & .590$\pm$.115 & .152$\pm$.097 & .557$\pm$.194 & .937$\pm$.050 & .488$\pm$.359 & .514$\pm$.318 \\
SG  & .492$\pm$.143 & .002$\pm$.002 & .508$\pm$.143 & .489$\pm$.182 & .026$\pm$.046 & .465$\pm$.095 & .760$\pm$.154 & .150$\pm$.163 & .289$\pm$.210 \\

VG  & .536$\pm$.147 & .126$\pm$.033 & .590$\pm$.142 & .543$\pm$.193 & .145$\pm$.067 & .602$\pm$.148 & .732$\pm$.152 & .157$\pm$.096 & .333$\pm$.176 \\
IG  & .494$\pm$.151 & .010$\pm$.012 & .514$\pm$.149 & \textbf{.596$\pm$.115} & \textbf{.166$\pm$.093} & .563$\pm$.194 & \textbf{.941$\pm$.046} & \textbf{.500$\pm$.356} & \textbf{.517$\pm$.315} \\
GDA & .541$\pm$.142 & .137$\pm$.037 & \textbf{.596$\pm$.146} & .542$\pm$.192 & .142$\pm$.069 & .600$\pm$.158 & .768$\pm$.135 & .178$\pm$.122 & .345$\pm$.201 \\
GIA & .533$\pm$.150 & .122$\pm$.036 & .589$\pm$.141 & .552$\pm$.177 & .154$\pm$.052 & .602$\pm$.166 & .772$\pm$.143 & .194$\pm$.141 & .363$\pm$.214 \\
SGA & .541$\pm$.142 & .137$\pm$.037 & \textbf{.596$\pm$.145} & .543$\pm$.192 & .143$\pm$.066 & .600$\pm$.159 & .770$\pm$.135 & .185$\pm$.132 & .353$\pm$.208 \\
SS  & \textbf{.542$\pm$.143} & \textbf{.137$\pm$.036} & \textbf{.596$\pm$.145} & .544$\pm$.192 & .148$\pm$.065 & \textbf{.604$\pm$.156} & .763$\pm$.137 & .182$\pm$.125 & .350$\pm$.201 \\
IGA & .532$\pm$.150 & .120$\pm$.036 & .587$\pm$.142 & .552$\pm$.179 & .154$\pm$.049 & .602$\pm$.164 & .777$\pm$.142 & .207$\pm$.152 & .369$\pm$.216 \\
Random  & .474$\pm$.146 & .001$\pm$.001 & .526$\pm$.146 & .483$\pm$.170 & .005$\pm$.005 & .518$\pm$.171 & .748$\pm$.141 & .046$\pm$.051 & .248$\pm$.137 \\
\bottomrule
\end{tabular}
}
\caption{EEG spatial-domain faithfulness scores. Features correspond to EEG channels. Values report mean $\pm$ standard deviation across EEG datasets and decoders. Best values in each column are in bold.}
\label{tab:full_eeg_spatial}
\end{table}

% ================= Temporal =================
\begin{table}[H]
\centering
\tiny
\setlength{\tabcolsep}{5pt}
\renewcommand{\arraystretch}{0.4} 
\resizebox{\textwidth}{!}{%
\begin{tabular}{l|ccc|ccc|ccc}
\toprule
& \multicolumn{3}{c|}{AIM}
& \multicolumn{3}{c|}{mdROAD}
& \multicolumn{3}{c}{Zeroing} \\
\textbf{Method}
& AOC & ABC & AUC
& AOC & ABC & AUC
& AOC & ABC & AUC \\
\midrule
GD  & .498$\pm$.133 & .007$\pm$.011 & .500$\pm$.131 & .405$\pm$.135 & .014$\pm$.015 & .598$\pm$.139 & .579$\pm$.092 & .187$\pm$.126 & .470$\pm$.112 \\
GI  & .511$\pm$.128 & .034$\pm$.033 & .519$\pm$.129 & .439$\pm$.123 & .077$\pm$.047 & .632$\pm$.144 & .650$\pm$.093 & .448$\pm$.215 & .513$\pm$.117 \\
SG  & .498$\pm$.133 & .007$\pm$.011 & .500$\pm$.130 & .405$\pm$.134 & .013$\pm$.015 & .598$\pm$.138 & .579$\pm$.088 & .188$\pm$.126 & .463$\pm$.110 \\

VG  & .526$\pm$.121 & .093$\pm$.053 & .567$\pm$.149 & .443$\pm$.135 & .099$\pm$.044 & .656$\pm$.133 & .601$\pm$.107 & .391$\pm$.169 & .551$\pm$.134 \\
IG  & .516$\pm$.127 & .045$\pm$.035 & .529$\pm$.131 & \textbf{.450$\pm$.128} & .095$\pm$.044 & .645$\pm$.136 & \textbf{.653$\pm$.089} & \textbf{.501$\pm$.223} & .542$\pm$.117 \\
GDA & .510$\pm$.129 & .048$\pm$.046 & .534$\pm$.137 & .445$\pm$.133 & .102$\pm$.045 & .657$\pm$.136 & .600$\pm$.107 & .395$\pm$.181 & .557$\pm$.137 \\
GIA & .519$\pm$.124 & .063$\pm$.044 & .545$\pm$.135 & .444$\pm$.133 & \textbf{.103$\pm$.047} & \textbf{.659$\pm$.136} & .595$\pm$.107 & .398$\pm$.173 & .558$\pm$.136 \\
SGA & .510$\pm$.130 & .048$\pm$.046 & .535$\pm$.136 & .445$\pm$.134 & .102$\pm$.047 & .657$\pm$.135 & .604$\pm$.095 & .395$\pm$.176 & .560$\pm$.140 \\
SS  & \textbf{.529$\pm$.120} & \textbf{.098$\pm$.055} & \textbf{.569$\pm$.149} & .445$\pm$.134 & \textbf{.103$\pm$.045} & .658$\pm$.135 & .603$\pm$.104 & .406$\pm$.179 & .560$\pm$.138 \\
IGA & .519$\pm$.123 & .068$\pm$.044 & .549$\pm$.137 & .443$\pm$.134 & .100$\pm$.047 & .657$\pm$.137 & .598$\pm$.107 & .395$\pm$.174 & \textbf{.562$\pm$.134} \\
Random  & .487$\pm$.134 & .003$\pm$.004 & .504$\pm$.132 & .394$\pm$.134 & .006$\pm$.005 & .599$\pm$.135 & .521$\pm$.103 & .126$\pm$.087 & .460$\pm$.103 \\
\bottomrule
\end{tabular}
}
\caption{EEG temporal-domain faithfulness scores. Features correspond to time windows. Values report mean $\pm$ standard deviation across EEG datasets and decoders. Best values in each column are in bold.}
\label{tab:full_eeg_temporal}
\end{table}

% ================= Spectral =================
\begin{table}[H]
\centering
\tiny
\setlength{\tabcolsep}{5pt}
\renewcommand{\arraystretch}{0.4} 
\resizebox{\textwidth}{!}{%
\begin{tabular}{l|ccc|ccc|ccc}
\toprule
& \multicolumn{3}{c|}{AIM}
& \multicolumn{3}{c|}{mdROAD}
& \multicolumn{3}{c}{Zeroing} \\
\textbf{Method}
& AOC & ABC & AUC
& AOC & ABC & AUC
& AOC & ABC & AUC \\
\midrule
GD  & \textbf{.602$\pm$.140} & \textbf{.258$\pm$.067} & \textbf{.656$\pm$.141} & \textbf{.523$\pm$.115} & .232$\pm$.039 & .709$\pm$.125 & \textbf{.822$\pm$.087} & .718$\pm$.157 & \textbf{.549$\pm$.093} \\
GI  & .523$\pm$.114 & .155$\pm$.106 & .620$\pm$.165 & .474$\pm$.112 & .175$\pm$.072 & .700$\pm$.129 & .661$\pm$.117 & .599$\pm$.140 & .545$\pm$.051 \\
SG  & .600$\pm$.139 & .250$\pm$.079 & .651$\pm$.147 & \textbf{.523$\pm$.116} & \textbf{.234$\pm$.038} & \textbf{.711$\pm$.124} & .821$\pm$.086 & \textbf{.735$\pm$.137} & .543$\pm$.101 \\
VG  & .501$\pm$.106 & .115$\pm$.113 & .578$\pm$.179 & .511$\pm$.128 & .186$\pm$.085 & .674$\pm$.143 & .608$\pm$.139 & .411$\pm$.205 & .459$\pm$.095 \\
IG  & .521$\pm$.113 & .154$\pm$.106 & .620$\pm$.166 & .475$\pm$.114 & .176$\pm$.073 & .701$\pm$.128 & .669$\pm$.109 & .577$\pm$.145 & .529$\pm$.063 \\
GDA & .514$\pm$.104 & .145$\pm$.117 & .615$\pm$.171 & .506$\pm$.118 & .203$\pm$.063 & .698$\pm$.133 & .631$\pm$.126 & .489$\pm$.191 & .516$\pm$.091 \\
GIA & .514$\pm$.100 & .128$\pm$.118 & .597$\pm$.178 & .499$\pm$.116 & .200$\pm$.065 & .701$\pm$.129 & .626$\pm$.124 & .500$\pm$.169 & .528$\pm$.080 \\
SGA & .516$\pm$.105 & .147$\pm$.116 & .616$\pm$.173 & .506$\pm$.117 & .206$\pm$.062 & .700$\pm$.132 & .636$\pm$.132 & .501$\pm$.191 & .520$\pm$.096 \\
SS  & .515$\pm$.105 & .152$\pm$.113 & .626$\pm$.165 & .505$\pm$.118 & .208$\pm$.066 & .702$\pm$.129 & .636$\pm$.128 & .509$\pm$.189 & .533$\pm$.089 \\
IGA & .517$\pm$.101 & .133$\pm$.115 & .601$\pm$.177 & .501$\pm$.116 & .203$\pm$.063 & .702$\pm$.128 & .631$\pm$.098 & .513$\pm$.164 & .536$\pm$.079 \\
Random  & .478$\pm$.136 & .003$\pm$.008 & .476$\pm$.097 & .400$\pm$.130 & .002$\pm$.002 & .519$\pm$.114 & .614$\pm$.034 & .069$\pm$.060 & .366$\pm$.073 \\
\bottomrule
\end{tabular}
}
\caption{EEG spectral-domain faithfulness scores. Features correspond to frequency components or bands. Values report mean $\pm$ standard deviation across EEG datasets and decoders. Best values in each column are in bold.}
\label{tab:full_eeg_spectral}
\end{table}

\subsection{Complete MoRF-LeRF Ranking Consistency Results}
\label{app:consistency_results}

\begin{table}[H]
\centering
\tiny
\setlength{\tabcolsep}{5pt}
\renewcommand{\arraystretch}{0.4} 
\resizebox{\textwidth}{!}{%
\begin{tabular}{c|ccc|ccc|ccc}
\toprule

& \multicolumn{3}{c|}{\textbf{OxfordPet}} 
& \multicolumn{3}{c|}{\textbf{ImageNet}} 
& \multicolumn{3}{c}{\textbf{BrainMRI}} \\

\cmidrule(lr){2-4} \cmidrule(lr){5-7} \cmidrule(lr){8-10}

\textbf{Masking}
& ResNet50 & EffNet-B0 & RepVGG-B0
& ResNet50 & EffNet-B0 & RepVGG-B0
& ResNet50 & EffNet-B0 & RepVGG-B0 \\

\midrule

AIM
& \textbf{.613$\pm$.321} & \textbf{.703$\pm$.256} & .477$\pm$.435
& .486$\pm$.541 & .554$\pm$.475 & .605$\pm$.465
& \textbf{.540$\pm$.249} & \textbf{.635$\pm$.325} & \textbf{.569$\pm$.152} \\

mdROAD
& .511$\pm$.321 & .622$\pm$.336 & \textbf{.600$\pm$.310}
& \textbf{.760$\pm$.294} & \textbf{.818$\pm$.187} & \textbf{.756$\pm$.201}
& .245$\pm$.373 & -.023$\pm$.399 & .528$\pm$.327 \\

Zeroing
& -.251$\pm$.246 & -.429$\pm$.407 & .090$\pm$.411
& -.620$\pm$.087 & -.549$\pm$.208 & -.313$\pm$.161
& .212$\pm$.397 & -.499$\pm$.470 & .355$\pm$.425 \\

\bottomrule
\end{tabular}
}
\caption{MoRF-LeRF ranking consistency for image models. Spearman correlations ($\rho$, mean $\pm$ standard deviation) quantify the agreement between explanation method faithfulness rankings obtained from complementary feature-masking orders. Best values across masking methods are in bold for each dataset--model pair.}
\label{tab:full_image_consistency}
\end{table}

\begin{table}[H]
\centering
\resizebox{\textwidth}{!}{
\begin{tabular}{l|cc|cc|cc|cc|cc|cc}
\toprule
& \multicolumn{4}{c|}{\textbf{AudioMNIST}} 
& \multicolumn{4}{c|}{\textbf{ESC50}} 
& \multicolumn{4}{c}{\textbf{MSOS}} \\
\cmidrule(lr){2-5} \cmidrule(lr){6-9} \cmidrule(lr){10-13}

& \multicolumn{2}{c|}{\textbf{Temporal}} & \multicolumn{2}{c|}{\textbf{Spatial}}
& \multicolumn{2}{c|}{\textbf{Temporal}} & \multicolumn{2}{c|}{\textbf{Spatial}}
& \multicolumn{2}{c|}{\textbf{Temporal}} & \multicolumn{2}{c}{\textbf{Spatial}} \\

\cmidrule(lr){2-3} \cmidrule(lr){4-5}
\cmidrule(lr){6-7} \cmidrule(lr){8-9}
\cmidrule(lr){10-11} \cmidrule(lr){12-13}

\textbf{Masking}
& \textbf{Res1d} & \textbf{AudioNet} & \textbf{AlexNet} & \textbf{CNN14}
& \textbf{Res1d} & \textbf{AudioNet} & \textbf{AlexNet} & \textbf{CNN14}
& \textbf{Res1d} & \textbf{AudioNet} & \textbf{AlexNet} & \textbf{CNN14} \\
\midrule

\textbf{AIM}
& \textbf{.754$\pm$.125} & .845$\pm$.121 & \textbf{.613$\pm$.263} & .782$\pm$.187
& \textbf{.801$\pm$.110} & .884$\pm$.126 & .765$\pm$.173 & \textbf{.846$\pm$.125}
& \textbf{.854$\pm$.092} & \textbf{.905$\pm$.090} & \textbf{.851$\pm$.123} & \textbf{.857$\pm$.082} \\

\textbf{mdROAD}
& .535$\pm$.215 & .933$\pm$.055 & .265$\pm$.381 & \textbf{.788$\pm$.073}
& .430$\pm$.179 & .886$\pm$.094 & \textbf{.744$\pm$.105} & .566$\pm$.274
& .362$\pm$.303 & .885$\pm$.107 & .832$\pm$.064 & .560$\pm$.208 \\

\textbf{Zeroing}
& .647$\pm$.229 & \textbf{.953$\pm$.024} & .412$\pm$.201 & .373$\pm$.186
& .505$\pm$.124 & \textbf{.896$\pm$.108} & .420$\pm$.269 & .539$\pm$.065
& .495$\pm$.238 & .876$\pm$.104 & -.120$\pm$.179 & .296$\pm$.322 \\

\bottomrule
\end{tabular}

}
\caption{MoRF-LeRF ranking consistency for audio models. Temporal domain corresponds to waveform models, and spatial/time--frequency domain corresponds to spectrogram models. Spearman correlations ($\rho$, mean $\pm$ standard deviation) quantify the stability of explanation method faithfulness rankings. Best values are in bold for each dataset--model pair.}
\label{tab:full_audio_consistency}
\end{table}

\begin{table}[H]
\centering
\tiny
\resizebox{\textwidth}{!}{
\begin{tabular}{ll|ccc|ccc|ccc}
\hline
\textbf{Framework} & \textbf{Domain} 
& \multicolumn{3}{c|}{\textbf{SMR}} 
& \multicolumn{3}{c|}{\textbf{ERN}} 
& \multicolumn{3}{c}{\textbf{SSVEP}} \\

 & 
 & EEGNet & ICNN & SCCNet 
 & EEGNet & ICNN & SCCNet 
 & EEGNet & ICNN & SCCNet \\
\hline

\multirow{3}{*}{Zero}
& Spatial
& -.206±.230 & .417±.145 & -.348±.419 & .581±.203 & .739±.175 & .647±.194 &  \textbf{.900±.061} & .830±.087 & .488±.242 \\

& Temporal 
& .477±.154 & .468±.184 & .587±.252 & .754±.223 &  \textbf{.748±.167} & .738±.205 &  \textbf{.698±.192} & .504±.387 &  \textbf{.502±.316} \\

& Spectral 
&  \textbf{.821±.098} & -.265±.584 & .041±.170 & -.045±.248 & .075±.364 & .145±.251 & .738±.146 & .368±.293 &  \textbf{.877±.069} \\
\hline

\multirow{3}{*}{mdROAD}
& Spatial 
& .136$\pm$.361 & .377$\pm$.280 & .557$\pm$.197 
& \textbf{.854$\pm$.098} & .583$\pm$.190 & .563$\pm$.203 
& .653$\pm$.199 & .733$\pm$.095 & .707$\pm$.179 \\

& Temporal 
& .548$\pm$.162 & .174$\pm$.267 & \textbf{.785$\pm$.149}
& .538$\pm$.162 & .700$\pm$.178 & .728$\pm$.164 
& .570$\pm$.227 & \textbf{.764$\pm$.141} & .353$\pm$.259 \\

& Spectral 
& .788$\pm$.120 & .101$\pm$.318 & \textbf{.362$\pm$.219}
& .383$\pm$.249 & .115$\pm$.244 & .166$\pm$.381 
& .318$\pm$.340 & .077$\pm$.324 & .023$\pm$.263 \\
\hline

\multirow{3}{*}{AIM}
& Spatial 
& \textbf{.830$\pm$.066} & \textbf{.816$\pm$.051} & \textbf{.821$\pm$.059}
& .767$\pm$.105 & \textbf{.773$\pm$.099} & \textbf{.833$\pm$.078}
& .881$\pm$.071 & \textbf{.870$\pm$.052} & \textbf{.862$\pm$.108} \\

& Temporal 
& \textbf{.921$\pm$.079} & \textbf{.790$\pm$.193} & .638$\pm$.130
& \textbf{.904$\pm$.096} & .614$\pm$.254 & \textbf{.805$\pm$.062}
& .434$\pm$.281 & .715$\pm$.071 & -.040$\pm$.370 \\

& Spectral 
& .813$\pm$.117 & \textbf{.755$\pm$.091} & .330$\pm$.218
& \textbf{.681$\pm$.226} & \textbf{.482$\pm$.207} & \textbf{.625$\pm$.113}
& \textbf{.755$\pm$.082} & \textbf{.592$\pm$.091} & .811$\pm$.061 \\

\hline
\end{tabular}
}
\caption{MoRF-LeRF ranking consistency for EEG models across masking methods, feature domains, datasets, and decoders. Spearman correlations ($\rho$, mean $\pm$ standard deviation) quantify the stability of explanation method faithfulness rankings. Best values are in bold within each dataset--model--domain configuration.}
\label{tab:full_eeg_consistency}
\end{table}

\subsection{Frequency-Correction Analysis for Unsigned Spectral Attribution}
\label{app:frequency_correction}
To further probe the effect of unsigned transformations in the spectral domain, we conduct a frequency-corrected masking experiment on the SSVEP dataset under the AIM framework. Instead of directly masking frequencies in $\Phi_f$, we apply perturbations to their corresponding half-frequency components:
\[
X'_{f/2} \leftarrow X^{\mathrm{adv}}_{f/2}, \quad \text{for } f \in \Phi_f.
\]
This correction improves faithfulness metrics under the MoRF setting, with gains in $\{\mathrm{AOC}, \mathrm{ABC}, \mathrm{AUC}\}$ across multiple unsigned methods, including GDA, GIA, SGA, and IGA. Because real EEG spectra contain nonlinear mixing and harmonic interactions, this correction should be viewed as a diagnostic analysis rather than a complete solution.

\subsection{Literature Map of Explanation Quality Evaluation}
\label{app:literature_table}
\begin{table}[H]
\centering
\scriptsize
\caption{Summary of explanation-quality evaluation studies. The table preserves the terminology used in prior work while grouping studies by their main evaluation motif. This work focuses on faithfulness evaluation and, more specifically, on the reliability of masking operators used to assess saliency maps.}
\label{tab:table1}
\begin{tabularx}{\textwidth}{p{2.7cm} p{3.3cm} p{2.7cm} X}
\toprule
\textbf{Evaluation motif} & \textbf{Terminology / representative studies} & \textbf{Removal or imputation} & \textbf{Main idea} \\
\midrule

Explanation axioms
& Sensitivity-$n$ \citep{ancona2017towards}; Completeness, Sensitivity, Linearity \citep{sundararajan2017axiomatic}; Summation to delta \citep{shrikumar2017learning}; Local Accuracy, Missingness, Consistency \citep{lundberg2017unified}; Sensitivity \citep{kindermans2019reliability}
& --
& Mathematical properties that attribution values should satisfy, usually independent of a specific masking operator. \\
\midrule

Explanation robustness
& Similarity \citep{adebayo2018sanity}; Sensitivity \citep{yeh2019fidelity}; Robustness and Sensitivity \citep{ravindran2023empirical}
& --; noise ratio
& Stability of attribution patterns under model randomization, input perturbation, noise, or similar generation settings. \\
\midrule

Removal-based faithfulness
& Quality \citep{samek2016evaluating}; Fidelity \citep{yeh2019fidelity, tomsett2020sanity, brocki2022fidelity}; Sensitivity \citep{cui2023towards}
& Remove
& Tests whether masking features identified as important leads to larger changes in model outputs. These methods are simple but can suffer from distribution shift. \\
\midrule

Retraining and debiased removal
& Importance Accuracy / ROAR \citep{hooker2018evaluating}; Fidelity / DiffROAR \citep{shah2021hiding}; Fidelity / ROAD \citep{rong2022consistent}; Reliability / ROAR \citep{torres2023evaluation}; Importance Accuracy / GOAR \citep{park2023geometric}; Effectiveness / corrupt-and-train \citep{turbe2023evaluation}
& ROAR, DiffROAR, ROAD, GOAR, corrupt-and-train
& Reduces masking-induced artifacts through retraining, interpolation, diffusion-based purification, or corruption-and-training strategies. \\
\midrule

Adversarial and OOD-aware faithfulness
& Quality / AR \citep{hsieh2020evaluations}; Faithfulness / OAR \citep{fang2024evaluating}; perturbation-based reliability \citep{nieradzik2025reliable}; F-Fidelity \citep{zheng2025ffidelity}
& AR, OAR, adversarial perturbation, fine-tuning with random masking
& Uses adversarial perturbation, OOD reweighting, or explanation-agnostic fine-tuning to reduce artifacts and improve reliability of faithfulness evaluation. \\
\midrule

Cross-modal evaluation
& EEG and time-series attribution evaluation \citep{apicella2022toward, cui2023towards, torres2023evaluation}; audio attribution evaluation \citep{becker2024audiomnist}; broader explainability evaluation discussions \citep{singh2021towards, rajpura2024explainable}; comparative metric studies \citep{gomez2022metrics, skliarov2025comparative}
& Remove, zero masking, ROAR, synthetic data, multiple metrics
& Highlights that evaluation results depend on feature definition, modality, masking rule, and metric choice. \\
\bottomrule
\end{tabularx}
\end{table}

\FloatBarrier
\clearpage

\subsection{Existing Assets and Licenses}
\label{app:assets_licenses}

We use existing benchmark datasets, model architectures, explanation methods, and open-source software packages only for research evaluation. We do not redistribute raw datasets, ImageNet images, or third-party pretrained checkpoints. All datasets and software packages are used according to their original licenses or terms of use. Table~\ref{tab:assets_licenses} summarizes the existing dataset assets used in this work.

\begin{table}[H]
\centering
\small
\caption{Existing dataset assets used in this work.}
\label{tab:assets_licenses}
\begin{tabular}{p{0.18\linewidth} p{0.18\linewidth} p{0.22\linewidth} p{0.30\linewidth}}
\toprule
\textbf{Asset} & \textbf{Type} & \textbf{Source / Citation} & \textbf{License / Terms of use} \\
\midrule
BCI Competition IV 2A / SMR 
& EEG dataset 
& Ref.~[46] 
& Creative Commons Attribution-NonCommercial-NoDerivatives 4.0 International License (CC BY-NC-ND 4.0); used for non-commercial research benchmarking. \\

BCI Challenge / ERN 
& EEG dataset 
& Ref.~[47] 
& Official competition / dataset terms; used for research benchmarking. \\

MAMEM SSVEP 
& EEG dataset 
& Ref.~[48] 
& Open Data Commons Attribution License v1.0 (ODC-By 1.0) for the PhysioNet release; used for research benchmarking. \\

AudioMNIST 
& Audio dataset 
& Ref.~[35] 
& MIT License. \\

ESC-50 
& Audio dataset 
& Ref.~[49] 
& Creative Commons Attribution-NonCommercial License; ESC-10 subset is distributed under Creative Commons Attribution License. \\

MSoS 
& Audio dataset 
& Ref.~[50] 
& Figshare / dataset-specific terms; released data files are used for research benchmarking under the original dataset terms. \\

Oxford-IIIT Pet 
& Image dataset 
& Ref.~[51] 
& Creative Commons Attribution-ShareAlike 4.0 International License (CC BY-SA 4.0). \\

ImageNet 
& Image dataset 
& Ref.~[52] 
& ImageNet terms of access; available for non-commercial research and educational purposes. We do not redistribute ImageNet images. \\

BrainMRI 
& Medical image dataset 
& Ref.~[53] 
& CC0: Public Domain license. \\

\bottomrule
\end{tabular}
\end{table}

\end{document}